\documentclass{article}

\usepackage{microtype}
\usepackage{graphicx}
\usepackage{subcaption}
\usepackage{booktabs} 

\usepackage{hyperref}

\usepackage[preprint]{icml2026}

\usepackage{amsmath}
\usepackage{amssymb}
\usepackage{mathtools}
\usepackage{amsthm}
\usepackage{multirow}
\usepackage[table]{xcolor}
\usepackage{enumitem}
\usepackage[capitalize,noabbrev]{cleveref}
\usepackage{bm}      

\theoremstyle{plain}

\theoremstyle{definition}

\theoremstyle{remark}

\usepackage[textsize=tiny]{todonotes}

\icmltitlerunning{Manifold-Aware Exploration for Reinforcement Learning in Video Generation}

\begin{document}

\twocolumn[
    \icmltitle{Manifold-Aware Exploration for Reinforcement Learning in Video Generation}



  \icmlsetsymbol{equal}{*}
  \icmlsetsymbol{projectlead}{$\ddagger$}
  \icmlsetsymbol{CorrespondingAuthors}{$\dagger$}

  \begin{icmlauthorlist}
    \icmlauthor{Mingzhe Zheng}{equal,sch,comp}
    \icmlauthor{Weijie Kong}{equal,comp}
    \icmlauthor{Yue Wu}{projectlead,comp}
    \icmlauthor{Dengyang Jiang}{sch}
    \icmlauthor{Yue Ma}{sch}
    \icmlauthor{Xuanhua He}{sch}
    \icmlauthor{Bin Lin}{comp}
    \icmlauthor{Kaixiong Gong}{comp}
    \icmlauthor{Zhao Zhong}{comp}
    \icmlauthor{Liefeng Bo}{comp}
    \icmlauthor{Qifeng Chen}{CorrespondingAuthors,sch}
    \icmlauthor{Harry Yang}{CorrespondingAuthors,sch}
  \end{icmlauthorlist}

  \icmlaffiliation{comp}{Tencent Hunyuan}
  \icmlaffiliation{sch}{HKUST}

  \icmlkeywords{Machine Learning, ICML}

  {%
\renewcommand\twocolumn[1][]{#1}%
\begin{center}
    \centering
    \vspace{5pt}
    \captionsetup{type=figure}
    \includegraphics[width=1.0\linewidth]{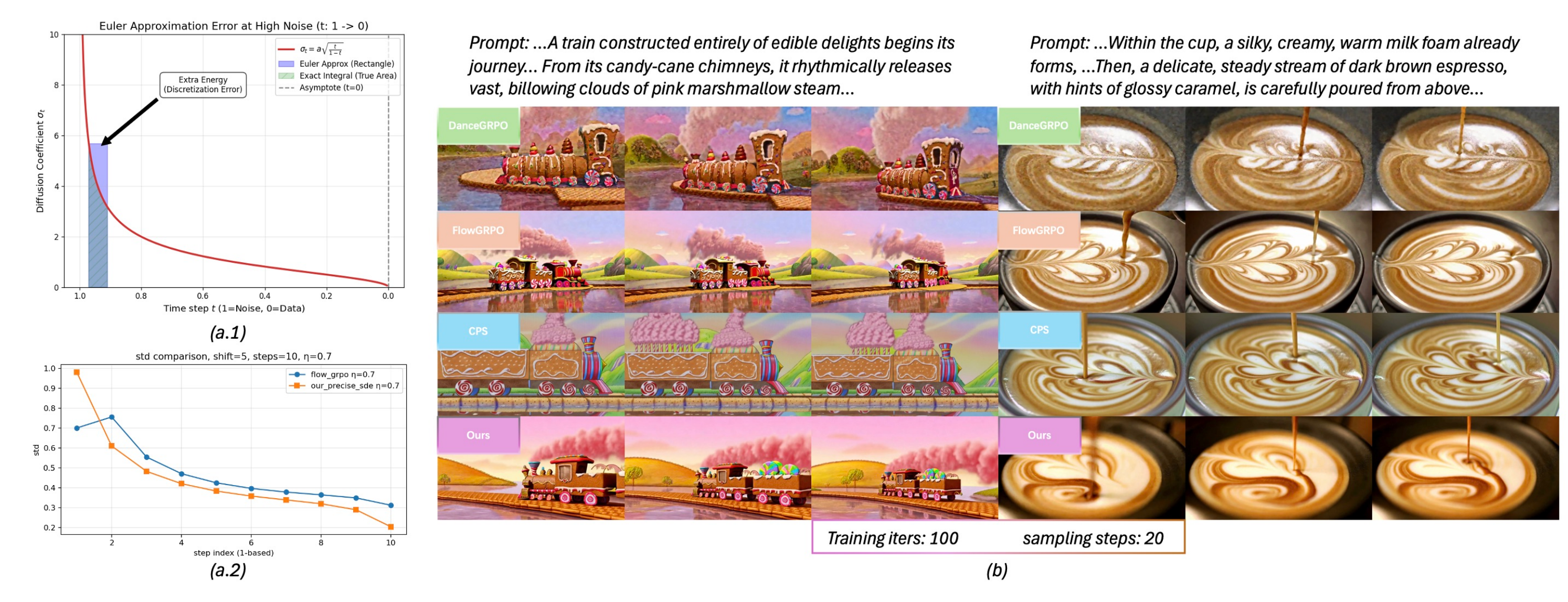}
    \vspace{-15pt}
     \caption{\textbf{Illustration of SAGE-GRPO.}
    \textbf{(Left)} (a.1) At a higher noise region, Euler-style discretization introduces a purple region of extra energy (discretization error) beyond the true integral; we focus on the true integral region below, not this extra energy. (a.2) Our precise SDE removes unnecessary noise energy in high-noise regions, enabling more precise exploration and a better-learned data manifold. \textbf{(Right)} (b) Our method with improved exploration yields more stable and better-aligned generations compared with DanceGRPO~\cite{dancegrpo}, FlowGRPO~\cite{flowgrpo}, and CPS~\cite{cps}.}
    \label{fig:teaser}
    \vspace{5pt}
    \centering
    
\end{center}%
}

]



\printAffiliationsAndNotice{\icmlWorkDone, \icmlEqualContribution, \icmlCorrespondingAuthors, \icmlProjectLeader}

\begin{abstract}
Group Relative Policy Optimization (GRPO) methods for video generation like FlowGRPO remain far less reliable than their counterparts for language models and images.
This gap arises because video generation has a complex solution space, and the ODE-to-SDE conversion used for exploration can inject excess noise, lowering rollout quality and making reward estimates less reliable, which destabilizes post-training alignment.
To address this problem, we view the pre-trained model as defining a valid video data manifold and formulate the core problem as constraining exploration within the vicinity of this manifold, ensuring that rollout quality is preserved and reward estimates remain reliable. 
We propose \textbf{SAGE-GRPO} (Stable Alignment via Exploration), which applies constraints at both micro and macro levels. 
At the micro level, we derive a \textit{precise manifold-aware SDE} with a logarithmic curvature correction and introduce a \textit{gradient norm equalizer} to stabilize sampling and updates across timesteps. 
At the macro level, we use a \textit{dual trust region} with a periodic moving anchor and stepwise constraints so that the trust region tracks checkpoints that are closer to the manifold and limits long-horizon drift. 
We evaluate SAGE-GRPO on HunyuanVideo1.5 using the original VideoAlign as the reward model and observe consistent gains over previous methods in VQ, MQ, TA, and visual metrics (CLIPScore, PickScore), demonstrating superior performance in both reward maximization and overall video quality.
The code and visual gallery are available at \href{https://dungeonmassster.github.io/SAGE-GRPO-Page/}{here}.
\end{abstract}
  
  \begin{figure}[t]
  \centering
  \includegraphics[width=0.48\textwidth]{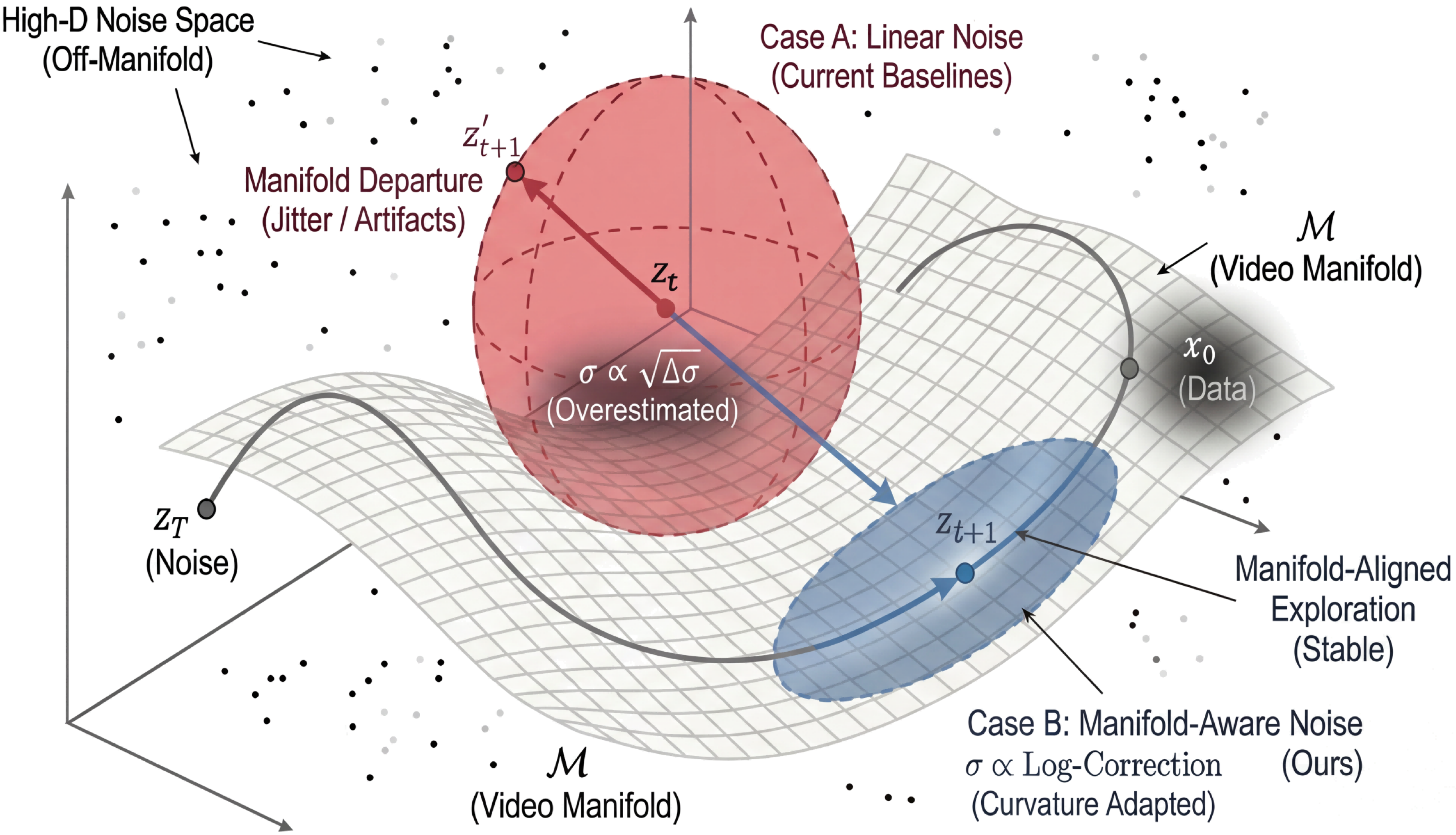} 
  \caption{\textbf{Geometric interpretation of noise injection strategies.}
    Conventional linear SDEs (red) inject exploration noise using first-order approximations, ignoring signal decay curvature and causing off-manifold drift that results in temporal jitter and artifacts. Our Manifold-Aware SDE (blue) uses a logarithmic correction term so that exploration noise is concentrated closer to the flow trajectory and the video manifold, reducing off-manifold drift.
        }
  \label{fig:2-manifold}
  \end{figure}
    
\section{Introduction}
\label{sec:intro}

Group Relative Policy Optimization (GRPO) is a direct way to align video generation models with reward signals~\cite{ddpm,scorebased,ddim,ma2025controllable,hunyuanvideo, hunyuanvideo1.5,wan,seedance}, but it has not yet been as reliable for video as it is for language models and images~\cite{deepseek_r1, deepseekmath, gpt, srpo}. 
In GRPO training for video generation, we must draw a group of rollouts by converting the deterministic ODE sampler into an SDE sampler so that the policy can explore through diverse samples~\cite{mixgrpo}. 
Video generation has a large, structured solution space, so this exploration is easily disturbed. 
Current video GRPO baselines such as DanceGRPO and FlowGRPO rely on an Euler-style discretization and first-order approximations when deriving the SDE noise standard deviation (as shown in Table~\ref{tab:sde_comparison})~\cite{ddpo,flowgrpo,dancegrpo}. 
The resulting first-order truncation error can inject excess noise energy during sampling (shown in Figure~\ref{fig:teaser}(a.1)), which lowers rollout quality in high-noise steps and makes reward evaluation less reliable. 
This raises the following question: \textit{how can we obtain an accurate sampling path that improves rollout quality and stabilizes GRPO for video generation?}

\begin{table}[t]
\centering
\caption{Comparison of SDE noise injection strategies used in video GRPO.}
\label{tab:sde_comparison}
\begin{tabular}{l|c}
\hline
\textbf{Method} & \textbf{Standard Deviation $\Sigma_t^{1/2}$} \\
\hline
DanceGRPO & $\eta \sqrt{\sigma_t - \sigma_{t+1}}$ \\
FlowGRPO & $\eta \sqrt{\frac{\sigma_t}{1-\sigma_t} (\sigma_t - \sigma_{t+1})}$ \\
Ours (Precise) & $\eta\sqrt{\left[-(\sigma_t - \sigma_{t+1}) + \log\left(\frac{1-\sigma_{t+1}}{1-\sigma_t}\right)\right]}$ \\
\hline
\end{tabular}
\end{table}

Flow-matching video generators parameterized by $\theta$ induce trajectories that are constrained by a pre-trained video generation model~\cite{rectifiedflow,flow-matching,rectifieddiffusion}. 
We treat this model as defining a valid data manifold $\mathcal{M} \subset \mathbb{R}^D$. Because the pre-trained parameters $\theta_0$ are not yet sufficient for the target reward, GRPO must update $\theta$ through exploration while keeping trajectories within the vicinity of $\mathcal{M}$ so that rollouts remain valid. 
As shown in Figure~\ref{fig:2-manifold}, FlowGRPO-style SDE exploration can overestimate the noise variance (red), push $z_t$ away from $\mathcal{M}$, and produce temporal jitter. 
We therefore define the core problem of GRPO for video generation as \textit{how to constrain exploration within the vicinity of the data manifold} so that each update improves rollouts while keeping reward evaluation reliable.

We propose \textbf{SAGE-GRPO} (Stable Alignment via Exploration), which organizes exploration at both micro and macro levels around the manifold. 
At the \textit{micro level}, we refine the discrete SDE and couple it with a \textit{gradient norm equalizer} as part of micro-scale exploration. 
Concretely, instead of using an area-based first-order variance approximation, we compute the noise variance by integrating diffusion coefficients over each step and add a logarithmic correction $\log\!\left(\frac{1-\sigma_{t+\Delta t}}{1-\sigma_t}\right)$, which yields a more accurate variance for ODE-to-SDE exploration. 
As in Figure~\ref{fig:teaser}(a.1), this corresponds to integrating only the effective energy under the curve rather than the extra discretization area, and Figure~\ref{fig:teaser}(a.2) shows that the resulting precise SDE uses smaller variance while staying closer to the underlying video manifold. 
Even with this corrected SDE, the diffusion process still has an inherent signal-to-noise imbalance across timesteps: gradients vanish at high noise ($t \to 1$) and explode at low noise ($t \to 0$), which biases learning toward certain phases. 
The Gradient Norm Equalizer normalizes optimization pressure across timesteps so that updates remain comparable in magnitude, which makes micro-level exploration more precise and stable.

With precise micro-level exploration, the policy after $N$ steps updates tends to move closer to the data manifold; periodically updating a reference model from this trajectory therefore creates a trust region centered at a more manifold-consistent policy. This reduces long-horizon drift and helps avoid off-manifold local optima, as suggested by the red region in Figure~\ref{fig:2-manifold}. 
Traditional Fixed KL constraints $D_{KL}(\pi_\theta || \pi_0)$ anchor the policy to the initial model $\pi_0$, but as training progresses the optimal policy $\pi^*$ may be far from $\pi_0$, which causes underfitting. 
Step-wise KL constraints $D_{KL}(\pi_\theta || \pi_{k-1})$ limit the magnitude of parameter updates per step (velocity control), ensuring smooth local transitions, but they only constrain the instantaneous update direction $\nabla_\theta$ and do not bound the cumulative displacement $\|\theta_k - \theta_0\|$ from the initial parameters. 
This allows unbounded drift: even if each step is small, the policy can move slowly but consistently away from the manifold over many steps, eventually leading to degradation or reward hacking. 
To counteract drift while preserving plasticity, we introduce a \textit{Periodical Moving Anchor} that updates the reference policy $\pi_{ref}$ every $N$ steps, creating a dynamic trust region that repeatedly recenters exploration near a manifold-consistent policy. 
We combine the moving anchor with step-wise constraints into a \textit{Dual Trust Region} objective that provides position control towards the manifold and velocity control between successive policies, forming a position-velocity controller that enables sustained plasticity.

We evaluate \textbf{SAGE-GRPO} on HunyuanVideo1.5~\cite{hunyuanvideo1.5} using the original VideoAlign evaluator~\cite{videoalign} (no reward-model fine-tuning) and observe consistent gains over baselines such as DanceGRPO~\cite{dancegrpo}, FlowGRPO~\cite{flowgrpo}, and CPS~\cite{cps} in both overall reward and temporal fidelity. 
Extensive ablations confirm that both the micro-level design (precise manifold-aware SDE with temporal gradient equalization) and the macro-level Dual Trust Region objective are necessary to reduce the stability–plasticity gap.

Our main contributions are as follows:
\vspace{-2mm}
\begin{itemize}[leftmargin=*,itemsep=0mm]
    \item We formulate GRPO for video generation as a manifold-constrained exploration problem and show that the ODE-to-SDE conversions used in existing methods can inject excess noise in high-noise steps, which reduces rollout quality and makes reward-guided updates less reliable.
    \item At the micro-level, we constrain exploration with a \textit{Precise Manifold-Aware SDE} and a \textit{Gradient Norm Equalizer}, so that sampling noise stays manifold-consistent and updates are balanced across timesteps.
    \item At the macro-level, we constrain long-horizon exploration with a \textit{Dual Trust Region} with moving anchors and step-wise constraints, so that the trust region tracks more manifold-consistent checkpoints and prevents drift.
\end{itemize}

\section{Related Work}
\label{sec:related_work}

\noindent \textbf{Reinforcement Learning for Diffusion and Flow Matching Models.}
Reinforcement learning has been adapted to fine-tune diffusion and flow matching models~\cite{flowgrpo,dancegrpo,imagereward,dmdr,diffusiondpo, xu2025scalar, lan2025flux, jin2025semantic, lin2025jarvisir, lin2025jarvisart, lin2025jarvisevo, zhang2026grpo} for alignment with human preferences. 
Early approaches such as DDPO~\cite{ddpo} and DPOK~\cite{dpok} treated the denoising process as a Markov Decision Process to enable policy gradient estimation.
Inspired by GRPO in language models~\cite{deepseekmath,deepseek_r1}, FlowGRPO~\cite{flowgrpo} and DanceGRPO~\cite{dancegrpo} adapted GRPO to visual generation via ODE-to-SDE conversion for stochastic exploration~\cite{mixgrpo}. 
However, existing methods rely on first-order noise approximations that can drive exploration off the data manifold and overlook the inherent gradient imbalance across timesteps.

\begin{figure*}[t]
\centering
\begin{subfigure}[b]{0.24\textwidth}
\centering
\includegraphics[width=\textwidth]{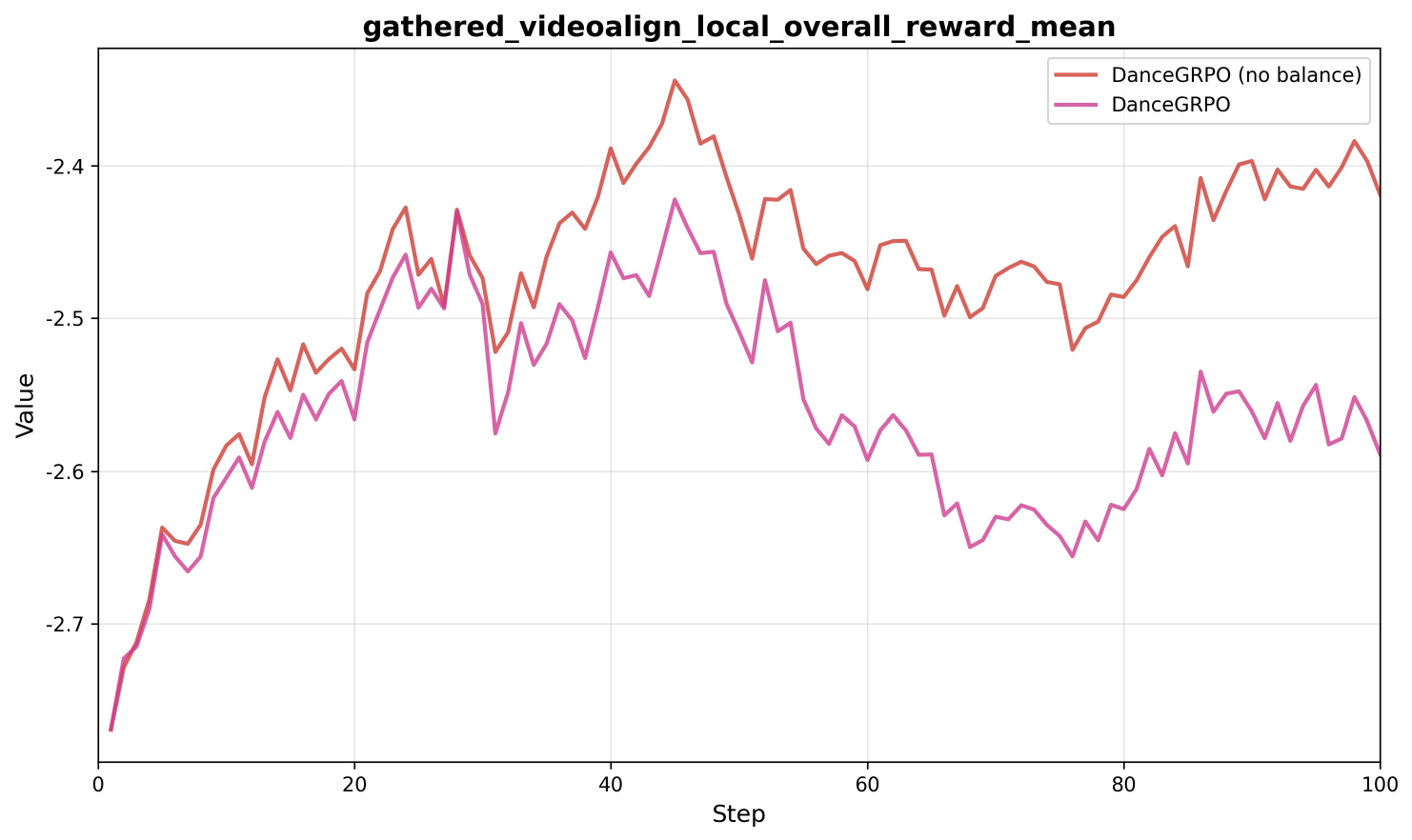}
\caption{DanceGRPO}
\label{fig:balance_dance}
\end{subfigure}
\hfill
\begin{subfigure}[b]{0.24\textwidth}
\centering
\includegraphics[width=\textwidth]{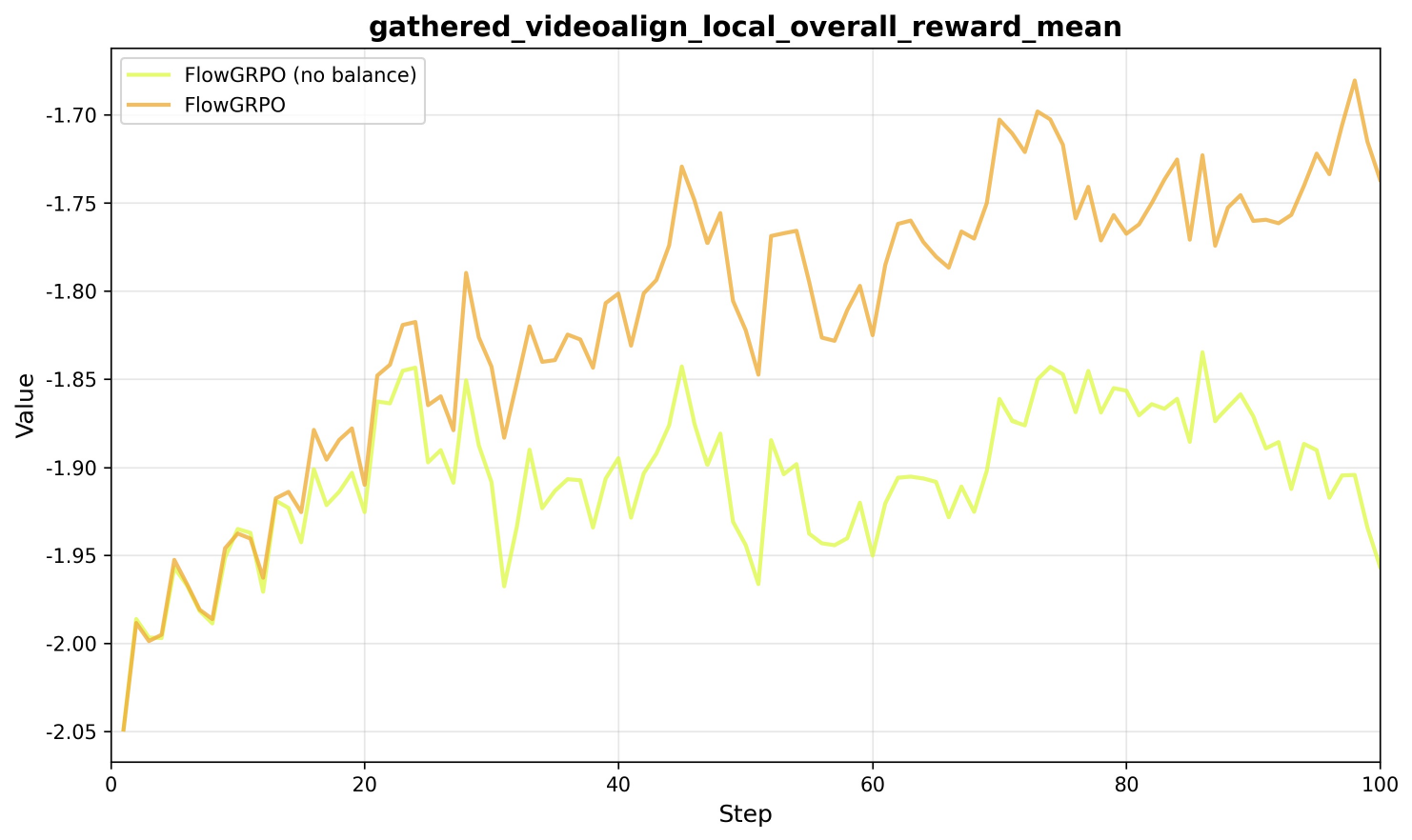}
\caption{FlowGRPO}
\label{fig:balance_flow}
\end{subfigure}
\hfill
\begin{subfigure}[b]{0.24\textwidth}
\centering
\includegraphics[width=\textwidth]{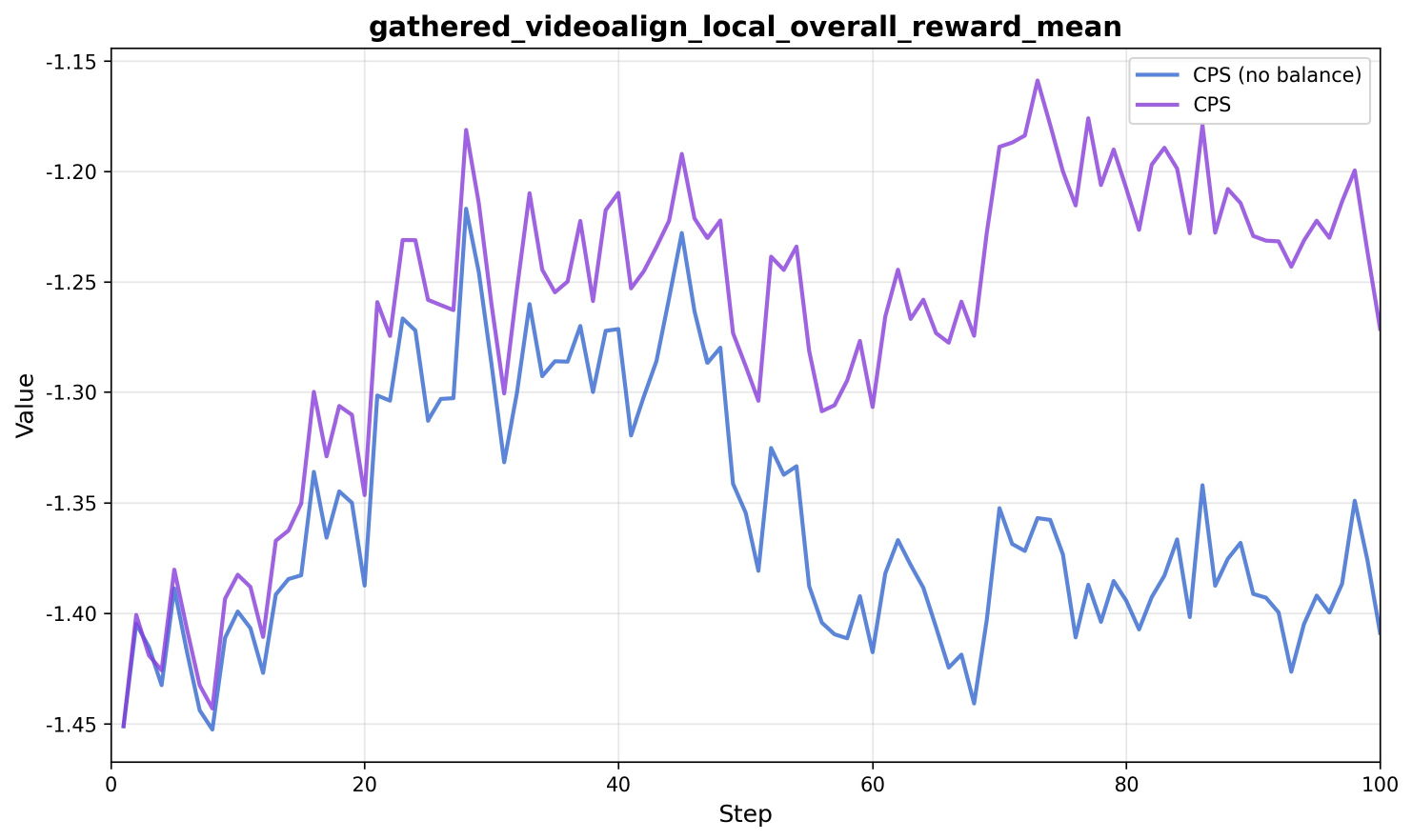}
\caption{CPS}
\label{fig:balance_cps}
\end{subfigure}
\hfill
\begin{subfigure}[b]{0.24\textwidth}
\centering
\includegraphics[width=\textwidth]{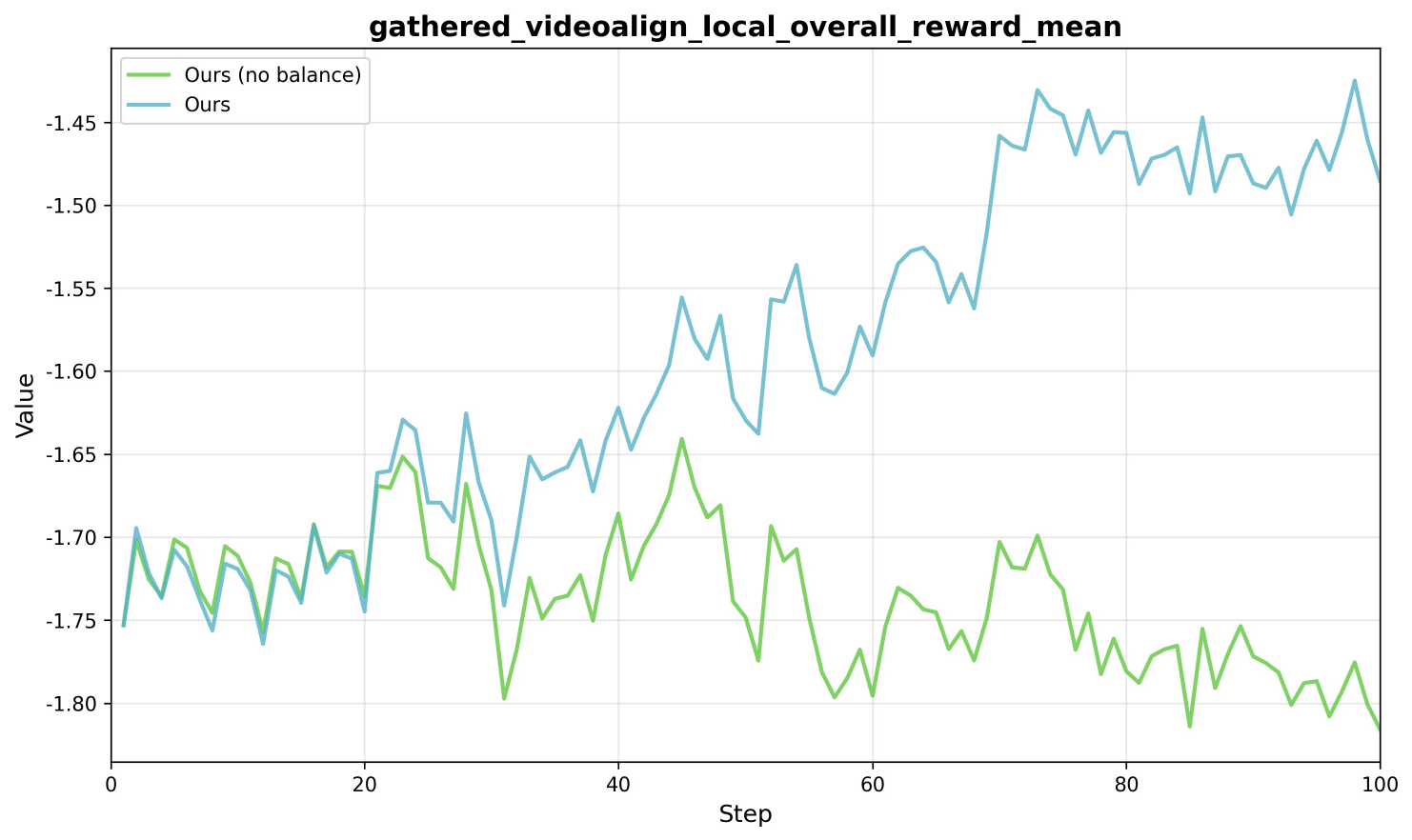}
\caption{Ours}
\label{fig:balance_ours}
\end{subfigure}
\caption{\textbf{Temporal gradient balancing ablation across SDE formulations.} 
Overall VideoAlign reward curves comparing runs with and without the Gradient Norm Equalizer. Without balancing, low-noise timesteps dominate optimization, leading to unstable or plateaued rewards. With balancing, reward curves become smoother with consistent improvement, and gradient scale variation is reduced from more than one order of magnitude to within a small constant factor.}
\label{fig:balance_ablation_main}
\end{figure*}

\noindent \textbf{Preference Alignment for Video Generation.}
Aligning video generation models with human preferences is an active research area~\cite{zheng2024videogen,long2025follow,videoreward,rewardforcing, he2025neighbor}. 
Building on video diffusion models~\cite{wan,hunyuanvideo,seedance}, researchers have developed video reward models~\cite{videoalign,visionreward,prfl, zhang2025diffusion} and alignment algorithms~\cite{li2024reward, gambashidze2024aligning, yu2024regularized, zhou2025fine, jia2025reward}. 
DanceGRPO~\cite{dancegrpo} extends image-based RL to video, while Self-paced GRPO~\cite{selfpacedgrpo} proposes curriculum learning that dynamically adjusts reward weights. 
However, current alignment frameworks face a stability-plasticity dilemma: strict constraints (e.g., fixed KL anchored to initialization) limit plasticity, while relaxed constraints trigger reward hacking or catastrophic forgetting~\cite{liu2025diversegrpo, li2025branchgrpo}. Unlike existing approaches that rely on heuristic scheduling or static anchors, our method integrates manifold-aware dynamics with a dual trust region to resolve this tension.

\begin{figure}[t]
\centering
\includegraphics[width=0.45\textwidth]{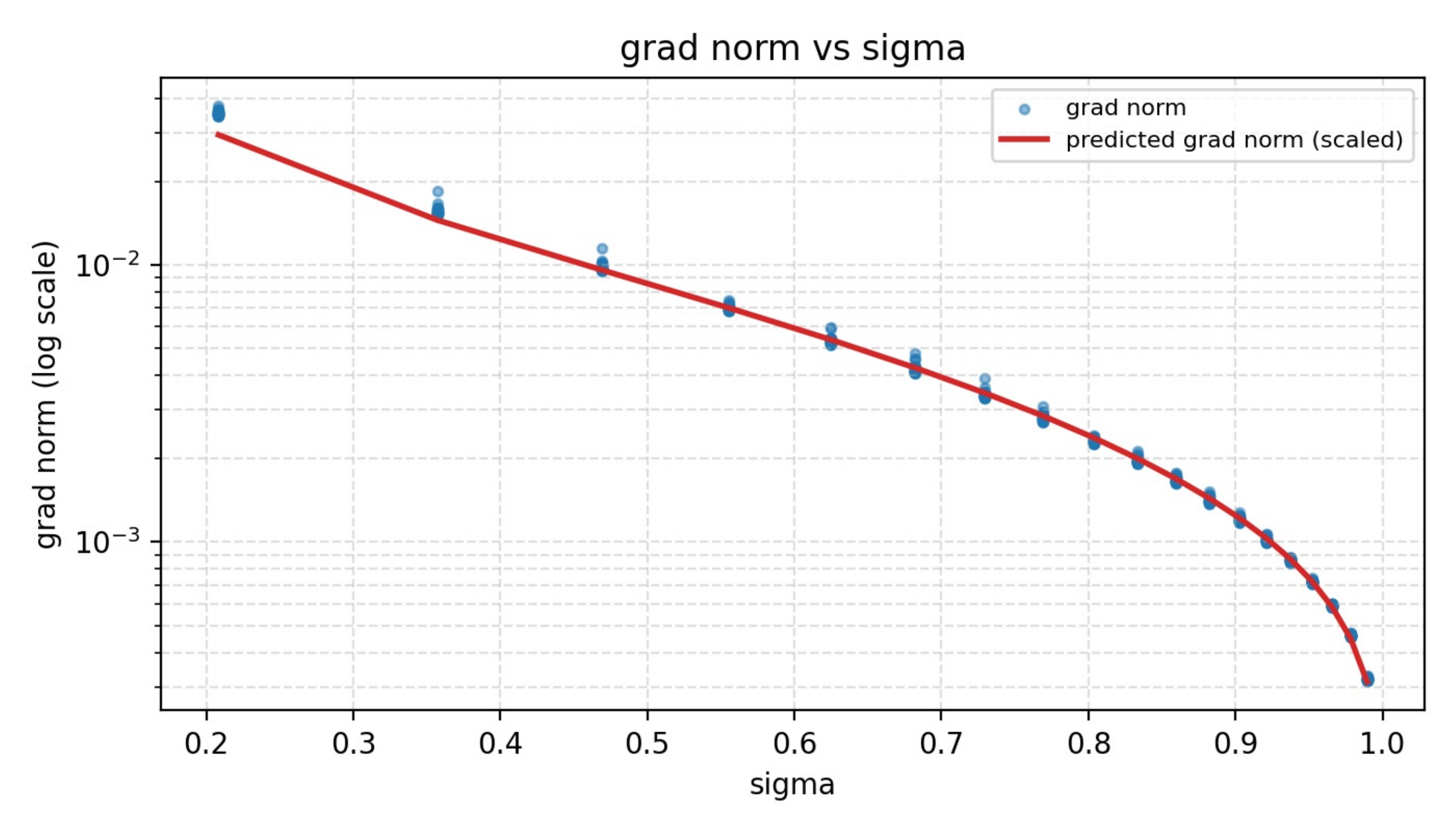}
\caption{Empirical gradient norm imbalance across noise levels. Observed norms (blue) decrease rapidly as $\sigma$ increases and match the predicted relationship (red) $\|\nabla \log \pi\| \propto 1/\Sigma_t^{1/2}$, leading to vanishing gradients at high noise ($\sigma \to 1$) and exploding gradients at low noise ($\sigma \to 0$).}
\label{fig:grad_norm_analysis}
\end{figure} 

\begin{figure*}[t]
    \centering
    \includegraphics[width=\textwidth]{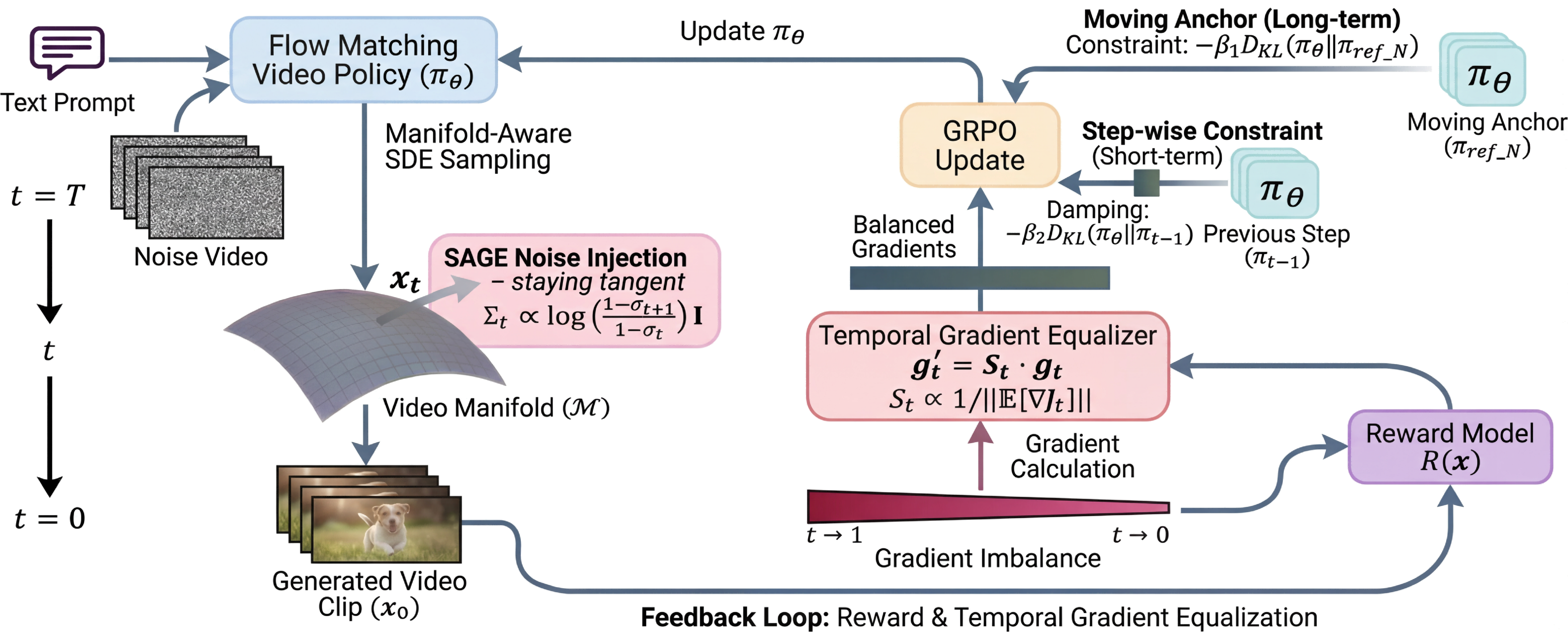}
    \caption{\textbf{The SAGE-GRPO Framework.} Our method resolves the stability-plasticity dilemma with three coupled components: \textbf{(Left)} a manifold-aware SDE that keeps exploration noise tangent to the video manifold, \textbf{(Middle)} a Temporal Gradient Equalizer that balances optimization across timesteps, and \textbf{(Right)} a Dual Trust Region that combines moving anchors and step-wise KL constraints for long-term stable alignment.}
    \label{fig:framework}
    
    \end{figure*}
    
    \section{Methodology}
    
    We formulate the problem of video alignment as maximizing the expected reward $J(\theta) = \mathbb{E}_{\mathbf{x}_0 \sim \pi_\theta}[R(\mathbf{x}_0)]$ within a Group Relative Policy Optimization (GRPO) framework. 
    However, a standard application of GRPO to video diffusion models faces specific challenges in maintaining stable and effective exploration on the video manifold. 
    \textbf{SAGE-GRPO} addresses these challenges by designing a unified exploration strategy that operates from micro-level noise injection to macro-level policy constraints, so that every exploration step remains valid and balanced across the diffusion process.
    
    \subsection{Preliminaries: Flow Matching and Group Relative Policy Optimization}
    
    \noindent \textbf{Flow Matching and Rectified Flow.}
    Flow Matching models generation as transport along a probability path $p_t(\mathbf{x})$ via an ordinary differential equation (ODE):
    \begin{equation}
        \frac{d\mathbf{x}_t}{dt} = \mathbf{v}_\theta(\mathbf{x}_t, t),
        \label{eq:flow_ode}
    \end{equation}
    where $\mathbf{v}_\theta$ is a neural velocity field. Rectified Flow uses the linear interpolation path:
    \begin{equation}
        \mathbf{x}_t = (1 - \sigma_t)\mathbf{x}_0 + \sigma_t \mathbf{z}_1,
        \label{eq:rectified_path}
    \end{equation}
    which implies the velocity field:
    \begin{equation}
        \mathbf{v}_\theta(\mathbf{x}_t, t) = \frac{d\mathbf{x}_t}{dt} = -\frac{d\sigma_t}{dt}(\mathbf{x}_0 - \mathbf{z}_1) = \frac{1}{1-\sigma_t}(\mathbf{x}_t - \mathbf{x}_0).
        \label{eq:velocity_field}
    \end{equation}
    \noindent \textbf{Group Relative Policy Optimization (GRPO).}
    Given a prompt $\mathbf{c}$, GRPO samples a group of $G$ rollouts and optimizes the diffusion policy $\pi_\theta$ using a group-normalized advantage:
    \begin{equation}
        \mathcal{L}_{GRPO}(\theta) = -\frac{1}{G} \sum_{i=1}^G A_i \cdot \sum_{t=1}^T \log \pi_\theta(\mathbf{x}_{t-1}^{(i)} | \mathbf{x}_t^{(i)}, \mathbf{c}),
        \label{eq:grpo_objective}
    \end{equation}
    where $T$ is the number of diffusion steps. We defer the reward composition, advantage normalization, and the stochastic rollout formulation to Appendix~\ref{sec:appendix} and only keep the key equations in the corresponding modules.
    
    \begin{figure*}[t]
        \centering
        \includegraphics[width=\textwidth]{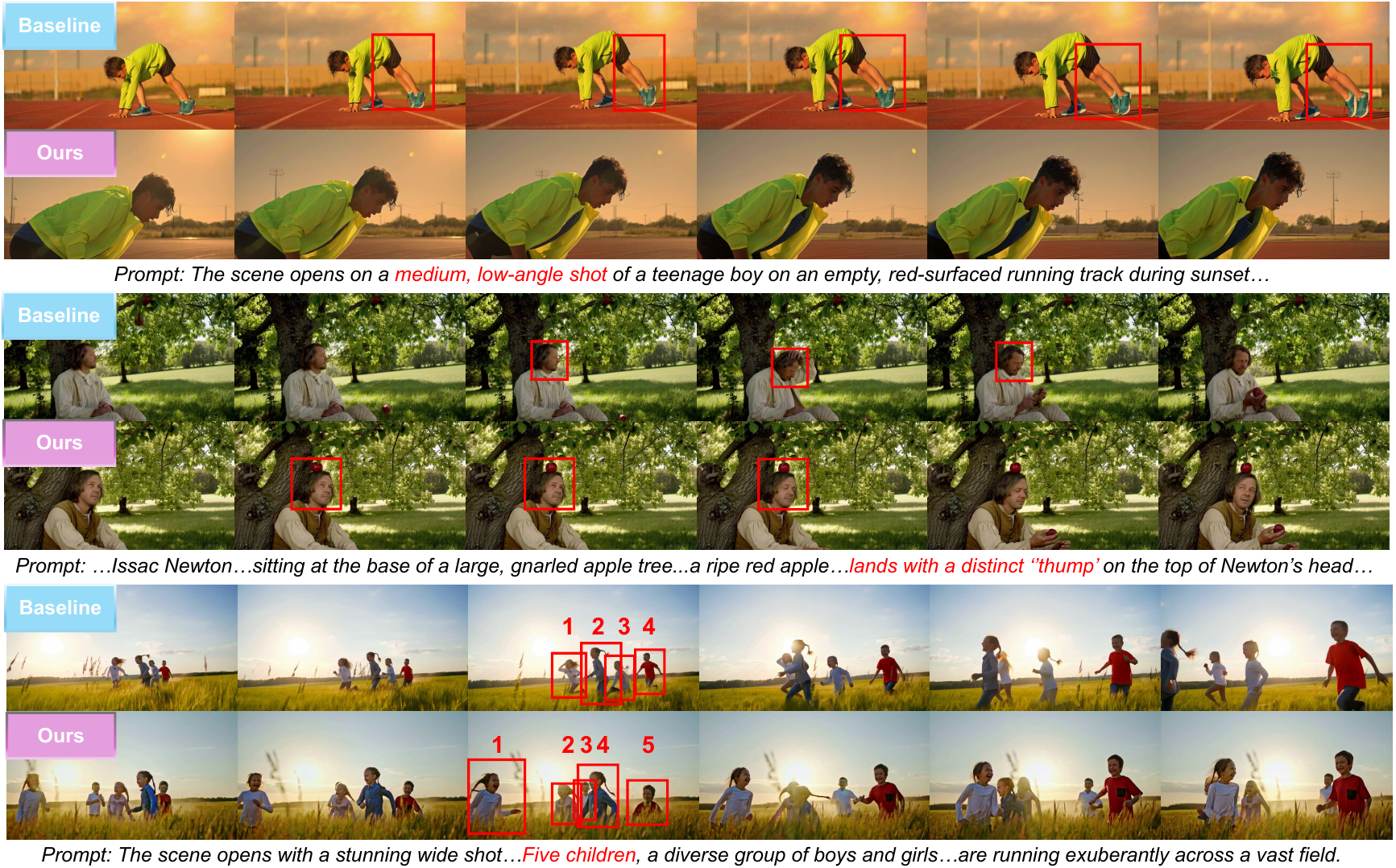}
        \caption{\textbf{Qualitative comparison against baselines.} Three prompts illustrate our core gains: (Top) Reduced temporal jitter while preserving accurate visual contents; (Middle) Enhanced alignment and photorealism under occlusion and lighting changes; (Bottom) Stronger semantic alignment with consistent prompt matching across frames.}
        \label{fig:qualitative_baseline}
    \end{figure*}
    
    \subsection{SAGE-GRPO Framework}
    
    \subsubsection{Micro-Level Exploration: Precise SDE and Gradient Equalization}
    \label{sec:method_micro}
    
    To enable stochastic exploration for GRPO, we perturb Rectified Flow with a marginal-preserving SDE whose noise stays aligned with the video manifold $\mathcal{M}\subset\mathbb{R}^D$ (Figure~\ref{fig:2-manifold}).
    The key challenge is computing the correct noise standard deviation $\Sigma_t^{1/2}$ during discrete SDE discretization.
    For a marginal-preserving SDE with diffusion coefficient $\varepsilon_t = \eta \sqrt{\sigma_t/(1-\sigma_t)}$, we integrate the variance over the interval $[\sigma_{t+1}, \sigma_t]$:
    \begin{equation}
        \Sigma_t = \int_{\sigma_{t+1}}^{\sigma_t} \varepsilon_s^2 \,\mathrm{d}s = \eta^2 \left[-(\sigma_t - \sigma_{t+1}) + \log\left(\frac{1-\sigma_{t+1}}{1-\sigma_t}\right)\right],
        \label{eq:precise_variance}
    \end{equation}
    where $\eta$ is the exploration scaling factor.
    The logarithmic term accounts for the geometric contraction of the signal coefficient $(1-\sigma_t)$, which linear approximations fail to capture.
    Taking the square root yields the noise standard deviation:
    \begin{equation}
        \Sigma_t^{1/2} = \eta \sqrt{-(\sigma_t - \sigma_{t+1}) + \log\left(\frac{1-\sigma_{t+1}}{1-\sigma_t}\right)}.
    \end{equation}
    
    Applying Euler-Maruyama discretization with timestep $\Delta t = \sigma_t - \sigma_{t+1}$:
    \begin{equation}
        \boxed{
        \mathbf{x}_{t+\Delta t} = \mathbf{x}_t + \mathbf{v}_\theta(\mathbf{x}_t, t)\Delta t + \frac{\Sigma_t}{2} \mathbf{s}_\theta(\mathbf{x}_t) + \Sigma_t^{1/2} \bm{\epsilon},
        }
        \label{eq:precise_sde}
    \end{equation}
    where $\boldsymbol{\epsilon} \sim \mathcal{N}(\mathbf{0}, \mathbf{I})$ injects stochasticity, $\mathbf{s}_\theta(\mathbf{x}_t) \approx -(\mathbf{x}_t - \hat{\mathbf{x}}_0)/\sigma_t^2$ is the score function estimate.
    Since $\Sigma_t$ the integrated variance is over $[\sigma_{t+1}, \sigma_t]$, the stochastic term is used $\Sigma_t^{1/2}$ directly without an additional $\sqrt{\Delta t}$ factor.
    The It\^{o} correction term $\frac{\Sigma_t}{2}\mathbf{s}_\theta(\mathbf{x}_t)$ ensures consistency with Rectified Flow marginals; a detailed derivation is provided in Appendix~\ref{app:sde_derivation}.
    
    
    As shown in Figure~\ref{fig:2-manifold}, our method creates a smaller, manifold-aligned exploration region (blue ellipsoid) that stays tangent to the flow trajectory, whereas conventional methods create larger, off-manifold exploration regions (red sphere) that cause state drift. 
    This geometric insight ensures that every exploration step remains within the legal video space, preventing temporal artifacts. 
    Even with correct noise injection, the diffusion process has an inherent signal-to-noise imbalance across timesteps: gradient norms vary by orders of magnitude (Figure~\ref{fig:grad_norm_analysis}), following a variance-gradient inverse relationship.
    For a Gaussian transition $\pi(\mathbf{x}_{t-1}|\mathbf{x}_t)=\mathcal{N}(\bm{\mu}_\theta,\Sigma_t\mathbf{I})$:
    \begin{equation}
        \| \nabla_{\bm{\mu}} \log \pi \| \propto \frac{1}{\Sigma_t^{1/2}},
        \label{eq:grad_norm_inverse}
    \end{equation}
    causing gradients to vanish at high noise ($t \to 1$) and explode at low noise ($t \to 0$), biasing learning toward certain phases.
    To counteract this imbalance, we estimate a per-timestep gradient scale $\mathcal{N}_t$ from the SDE parameters (Appendix~\ref{app:equalizer_details}) and apply a robust normalization:
    \begin{equation}
        \boxed{
        S_t = \frac{\text{Median}(\{\mathcal{N}_\tau\}_{\tau=1}^T)}{\mathcal{N}_t + \epsilon},
        }
        \label{eq:grad_equalizer}
    \end{equation}
    where $\epsilon$ is a small constant.
    This equalization normalizes optimization pressure across timesteps so that structural and textural updates contribute equally; empirical validation is provided in Figure~\ref{fig:balance_ablation_main} and Appendix~\ref{app:equalizer_details}.
    
    \noindent \textbf{GRPO With Composite Reward and Group-Normalized Advantage.}
    We score each rollout $\mathbf{x}_0$ by a composite reward $R(\mathbf{x}_0)$ and compute the group-normalized advantage $A_i$:
    \begin{equation}
        A_i = \frac{r_i - \mu_R}{\sigma_R + \epsilon},
        \label{eq:reward_adv_main}
    \end{equation}
    where $r_i=R(\mathbf{x}_0^{(i)})$, $\mu_R=\frac{1}{G}\sum_{j=1}^G r_j$, and $\sigma_R^2=\frac{1}{G}\sum_{j=1}^G (r_j-\mu_R)^2$. Full definitions and implementation-aligned details are in Appendix~\ref{app:grpo_details}.

\begin{table*}[t]
    \caption{\textbf{Main Comparison on Video Generation Benchmarks.} Comparison of SAGE-GRPO with baselines under two reward settings. The first row reports the original HunyuanVideo 1.5 performance. For each method, we report results \textit{without} KL regularization (\textbf{w/o KL}) and \textit{with} their Fixed KL constraints (\textbf{w/ Fixed KL}). For SAGE-GRPO, we demonstrate the \textbf{w/ Dual Moving KL} mechanism. \textbf{Bold}, \underline{underline}, and \textcolor{gray}{gray} indicate the best, second best, and third best results, computed across both settings (A+B).}
    \vspace{-12pt}
    \label{tab:main_results}
    \begin{center}
    \begin{tabular}{l|l|llll|ll}
    \hline
    \multirow{2}{*}{\bf Method} & \multirow{2}{*}{\bf Configuration} & \multicolumn{4}{c|}{\bf VideoAlign Metrics} & \multicolumn{2}{c}{\bf Visual Metrics} \\
     & & \multicolumn{1}{l}{\bf Overall} & \multicolumn{1}{l}{\bf VQ} & \multicolumn{1}{l}{\bf MQ} & \multicolumn{1}{l|}{\bf TA} & \multicolumn{1}{l}{\bf CLIPScore} & \multicolumn{1}{l}{\bf PickScore} \\
    \hline
    HunyuanVideo 1.5 (Original) & - & 0.0654 & -0.7539 & -0.5870 & 1.4063 & 0.5409 & 0.7397 \\
    \hline
    \multicolumn{8}{c}{\cellcolor{blue!5} \bf Setting A: Averaged Rewards ($w_{vq}=1.0, w_{mq}=1.0, w_{ta}=1.0$)} \\
    \hline
    \multirow{2}{*}{DanceGRPO} 
     & w/o KL & 0.2768 & -0.7589 & \underline{-0.3852} & 1.4209 & 0.5386 & 0.7378 \\
     & w/ Fixed KL & 0.0979 & -0.8077 & -0.5091 & 1.4147 & 0.5403 & 0.7355 \\
    \hline
    \multirow{2}{*}{FlowGRPO}
     & w/o KL & 0.2733 & -0.7151 & -0.5286 & 1.5170 & 0.5443 & 0.7394 \\
     & w/ Fixed KL & 0.1880 & -0.6771 & -0.5912 & 1.4563 & 0.5431 & 0.7407 \\
    \hline
    \multirow{2}{*}{CPS}
     & w/o KL & \underline{0.6343} & \underline{-0.4855} & \textcolor{gray}{-0.4021} & \underline{1.5219} & \underline{0.5479} & \textcolor{gray}{0.7412} \\
     & w/ Fixed KL & 0.0928 & -0.7156 & -0.5825 & 1.3908 & \underline{0.5479} & 0.7369 \\
    \hline
    \multirow{3}{*}{\textbf{SAGE-GRPO}}
     & w/o KL & \textcolor{gray}{0.4859} & -0.6104 & -0.4141 & 1.5104 & 0.5423 & 0.7360 \\
     & w/ Fixed KL & 0.2244 & -0.7438 & -0.5320 & 1.5001 & 0.5446 & 0.7382 \\
     & w/ Dual Mov KL & 0.2173 & -0.7881 & -0.4249 & 1.4303 & 0.5430 & \textbf{0.7452} \\
    \hline
    \multicolumn{8}{c}{\cellcolor{blue!5} \bf Setting B: Alignment-Focused ($w_{vq}=0.5, w_{mq}=0.5, w_{ta}=1.0$)} \\
    \hline
    \multirow{2}{*}{DanceGRPO}
     & w/o KL & -0.2172 & -0.8854 & -0.6218 & 1.2901 & 0.5439 & 0.7352 \\
     & w/ Fixed KL & 0.1290 & -0.7739 & -0.5083 & 1.4112 & 0.5452 & 0.7276 \\
    \hline
    \multirow{2}{*}{FlowGRPO}
     & w/o KL & 0.4773 & \textcolor{gray}{-0.5671} & -0.4731 & 1.5175 & 0.5403 & 0.7349 \\
     & w/ Fixed KL & 0.2103 & -0.6654 & -0.5506 & 1.4263 & 0.5427 & 0.7408 \\
    \hline
    \multirow{2}{*}{CPS}
     & w/o KL & 0.3694 & -0.6650 & -0.5325 & \textbf{1.5669} & \underline{0.5479} & 0.7311 \\
     & w/ Fixed KL & 0.3705 & -0.6121 & -0.4787 & 1.4613 & \textcolor{gray}{0.5458} & 0.7364 \\
    \hline
    \multirow{3}{*}{\textbf{SAGE-GRPO}}
     & w/o KL & -0.1222 & -0.8720 & -0.6046 & 1.3544 & 0.5404 & 0.7357 \\
     & w/ Fixed KL & 0.2857 & -0.7062 & -0.4425 & 1.4344 & 0.5414 & 0.7377 \\
     & \textbf{w/ Dual Mov KL} & \textbf{0.8066} & \textbf{-0.4765} & \textbf{-0.2384} & \textcolor{gray}{1.5216} & \textbf{0.5484} & \underline{0.7420} \\
    \hline
    \end{tabular}
    \end{center}
    \end{table*}
    
    \begin{figure*}[t]
    \centering
    \begin{subfigure}[b]{0.32\textwidth}
    \centering
    \includegraphics[width=\textwidth]{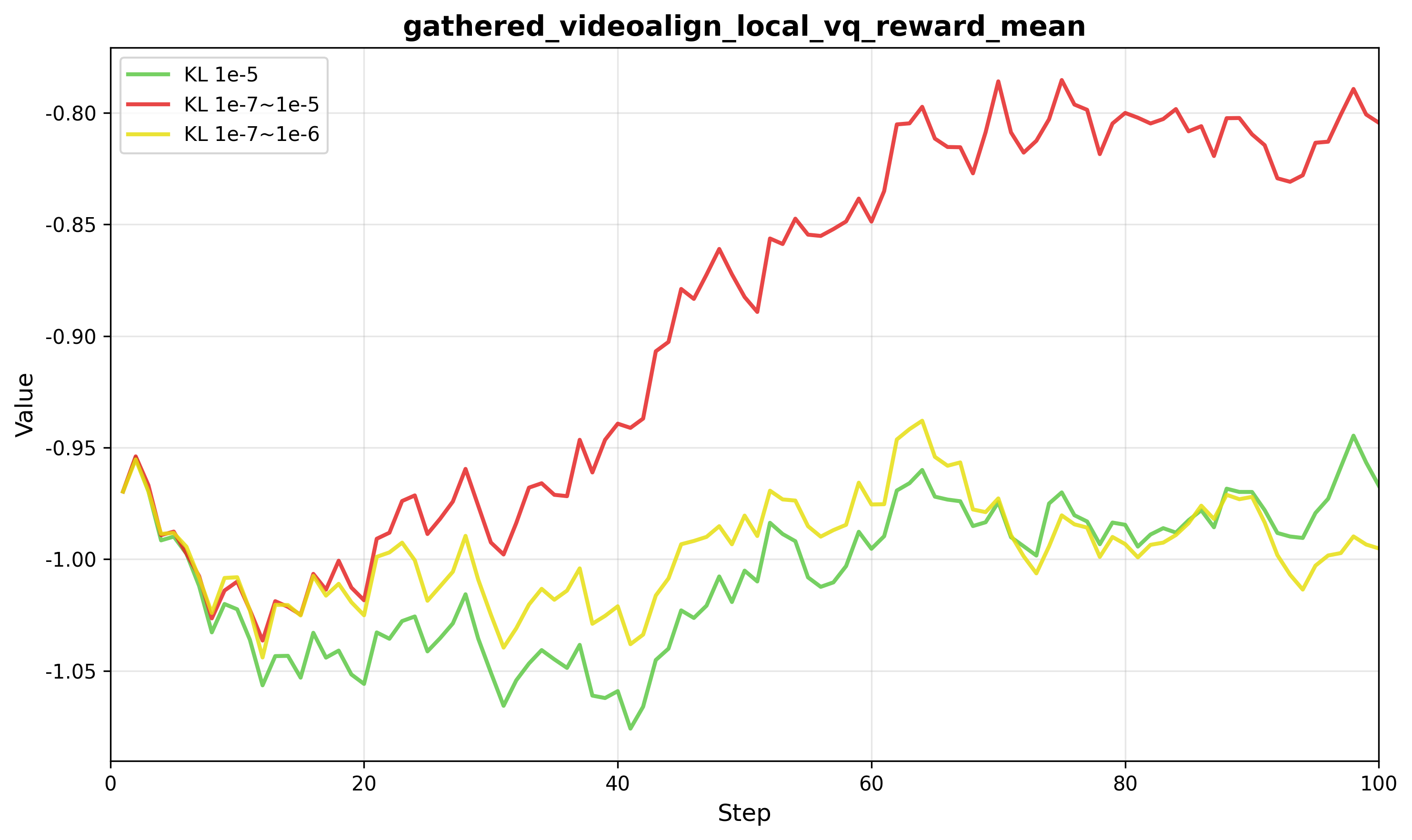}
    \caption{VQ reward}
    \label{fig:kl_weight_vq}
    \end{subfigure}
    \hfill
    \begin{subfigure}[b]{0.32\textwidth}
    \centering
    \includegraphics[width=\textwidth]{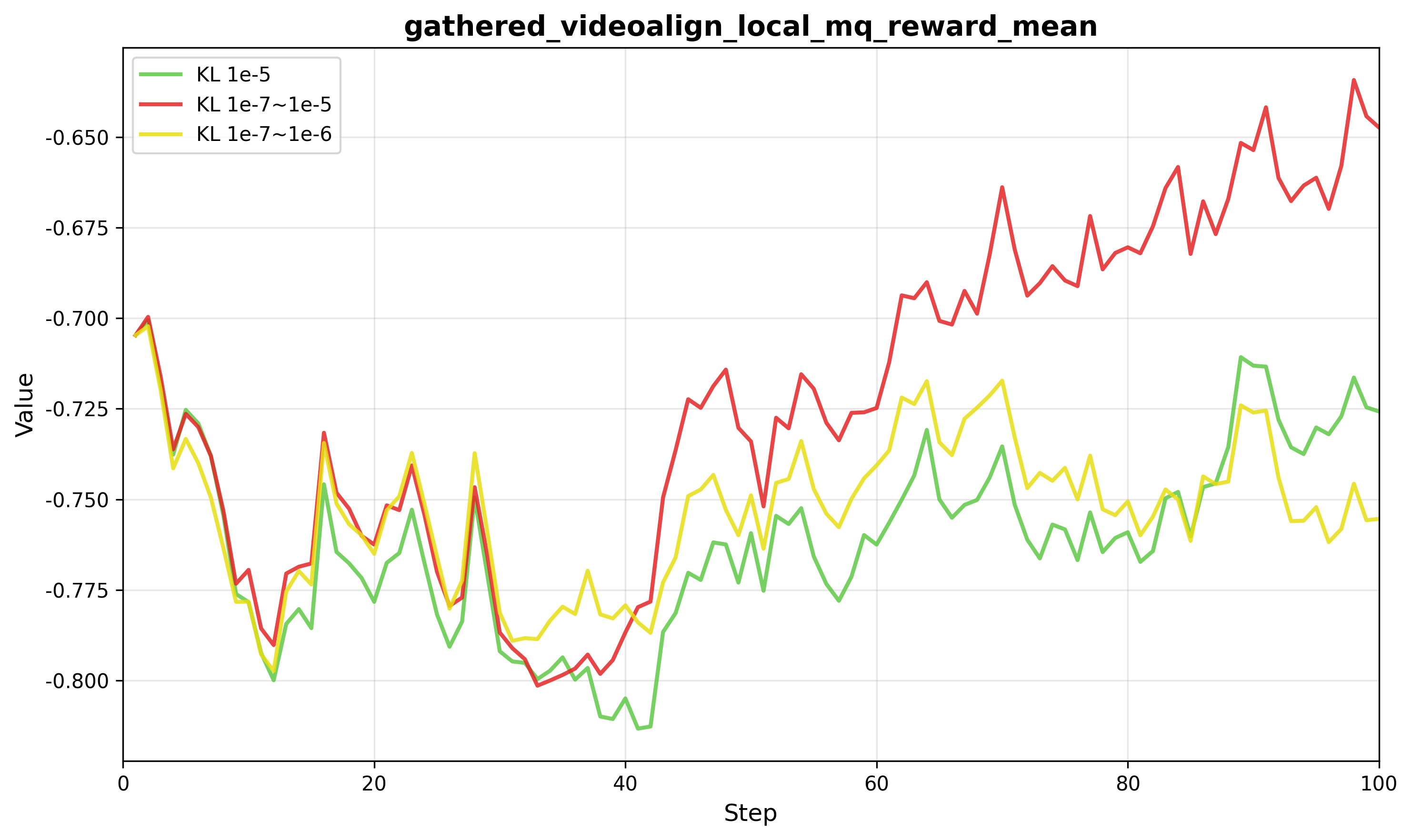}
    \caption{MQ reward}
    \label{fig:kl_weight_mq}
    \end{subfigure}
    \hfill
    \begin{subfigure}[b]{0.32\textwidth}
    \centering
    \includegraphics[width=\textwidth]{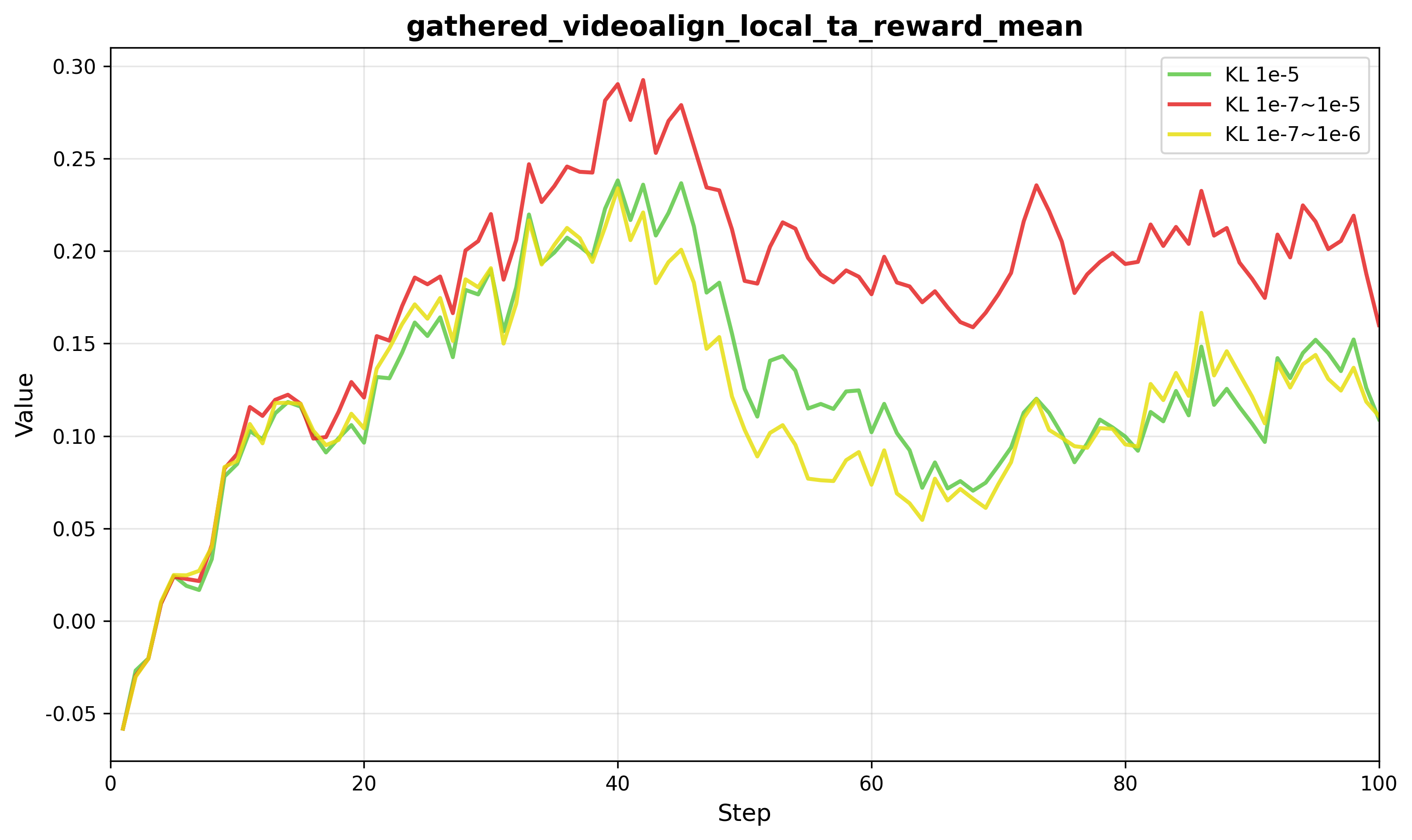}
    \caption{TA reward}
    \label{fig:kl_weight_ta}
    \end{subfigure}
    \caption{\textbf{KL weight ablation on VideoAlign rewards.} 
    Comparison of three KL weight schedules: fixed $10^{-5}$ (green), two-stage $10^{-7} \rightarrow 10^{-5}$ (red), and two-stage $10^{-7} \rightarrow 10^{-6}$ (yellow). The two-stage schedule $10^{-7} \rightarrow 10^{-5}$ achieves the strongest and most consistent gains across VQ, MQ, and TA, supporting gradually increasing $\lambda_{KL}$ to tighten the trust region (Appendix~\ref{app:adaptive_kl}).}
    \label{fig:kl_weight_ablation}
    \end{figure*}
    
    \subsubsection{Macro-Level Exploration: Dual Trust Region Optimization}
    \label{sec:method_dual_kl}
    
    With micro-level exploration stabilized, we aim to prevent the policy model from drifting away from the data manifold and getting stuck in off-manifold local optima (Figure~\ref{fig:2-manifold}).
    We frame KL divergence as a dynamic anchoring mechanism that constrains exploration towards the data manifold.

    \noindent \textbf{KL Divergence as Dynamic Anchor.} 
    For a Gaussian policy $\pi(\mathbf{x}_{t-1}|\mathbf{x}_t) = \mathcal{N}(\bm{\mu}_\theta, \Sigma_t \mathbf{I})$, the KL divergence between the current policy $\pi_\theta$ and a reference policy $\pi_{ref}$ is:
    \begin{equation}
        D_{KL}(\pi_\theta || \pi_{ref}) = \mathbb{E}_{\mathbf{x}_t \sim \pi_\theta} \left[ \frac{(\bm{\mu}_\theta - \bm{\mu}_{ref})^2}{2\Sigma_t^2} \right] \approx \frac{(\bm{\mu}_\theta - \bm{\mu}_{ref})^2}{2\Sigma_t^2},
        \label{eq:kl_divergence}
    \end{equation}
    where $\bm{\mu}_\theta$ and $\bm{\mu}_{ref}$ are the mean predictions of the current and reference policies, respectively. 
    KL divergence acts as a distance metric in policy space, anchoring the current policy to the reference. 
    The choice of reference determines the constraint nature: a fixed reference creates a hard constraint, while a moving reference enables adaptive exploration.
    
    \noindent \textbf{Fixed KL: Hard Constraint Limiting Optimality.} 
    Traditional approaches use a fixed reference policy $\pi_{ref} = \pi_0$ from the pretrained video generation model. 
    The constraint $D_{KL}(\pi_\theta || \pi_0)$ forces the policy to remain close to the initial distribution. 
    However, as training progresses, the optimal policy $\pi^*$ may be far from $\pi_0$, and forcing $D_{KL}(\pi_\theta || \pi_0)$ to be small prevents reaching $\pi^*$, leading to underfitting, which is too restrictive for long-term optimization where the policy needs to explore regions far from initialization.
    
    \noindent \textbf{Step-wise KL: Velocity Constraint.} 
    Step-wise KL uses the previous step's policy as reference: $\pi_{ref} = \pi_{k-1}$, where $k$ denotes the optimization step. 
    This constraint $D_{KL}(\pi_\theta || \pi_{k-1})$ acts as a velocity limit, restricting the magnitude of parameter updates per step:
    \begin{equation}
        \|\nabla_\theta D_{KL}(\pi_\theta || \pi_{k-1})\| \propto \|\bm{\mu}_\theta - \bm{\mu}_{k-1}\| / \Sigma_t,
        \label{eq:stepwise_kl}
    \end{equation}
    ensuring smooth local transitions. 
    However, velocity control alone only limits the magnitude of $\nabla_\theta$ (the update direction) but does not bound the cumulative displacement $\|\theta_k - \theta_0\|$ from the initial parameters. 
    This allows unbounded drift: the policy move slowly but consistently away from the manifold, eventually leading to degradation or reward hacking.
    
    \noindent \textbf{Periodical Moving KL: Position Control via Dynamic Trust Region.} 
    To counteract drift while maintaining plasticity, we introduce \textbf{Periodical Moving KL} that uses a periodically updated reference policy $\pi_{ref} = \pi_{k-N}$, where $N$ is the update interval. 
    For every $N$ optimization step, we update the reference model: $\pi_{ref} \leftarrow \pi_\theta$, creating a resetting anchor mechanism. 
    This allows the model to perform local exploration within $N$ steps, then establish the new position as a safe region:
    \begin{equation}
        D_{KL}(\pi_\theta || \pi_{ref\_N}) = \frac{(\bm{\mu}_\theta - \bm{\mu}_{ref\_N})^2}{2\Sigma_t^2},
        \label{eq:moving_kl}
    \end{equation}
    where $\bm{\mu}_{ref\_N}$ is the mean prediction from the reference model updated $N$ steps ago.
    This creates a dynamic trust region that periodically resets the safe zone, similar to a multi-stage relaxed version of TRPO~\cite{trpo}, enabling the model to climb the reward landscape in stages (plasticity) while tethered to a valid distribution (stability).
    
    \noindent \textbf{Dual KL: Position-Velocity Controller.} 
    We combine these two mechanisms into a \textbf{dual KL} objective that provides both position and velocity control:
    \begin{equation}
        \mathcal{L}_{KL} = \beta_{pos} \cdot D_{KL}(\pi_\theta || \pi_{ref\_N}) + \beta_{vel} \cdot D_{KL}(\pi_\theta || \pi_{k-1}),
        \label{eq:dual_kl}
    \end{equation}
    where $\beta_{pos}$ and $\beta_{vel}$ are weighting coefficients.
    The position term $D_{KL}(\pi_\theta || \pi_{ref\_N})$ provides the primary directional anchor, preventing long-term drift by constraining the policy to remain within a reasonable distance from a recent valid distribution. 
    The velocity term $D_{KL}(\pi_\theta || \pi_{k-1})$ acts as a damping factor, smoothing instantaneous updates and preventing abrupt policy changes.
    In practice, we compute the step-wise KL using log-probability differences from the rollout phase:
    \begin{equation}
        D_{KL}(\pi_\theta || \pi_{k-1}) \approx \mathbb{E}[\log \pi_{k-1}(\mathbf{x}_{t-1}|\mathbf{x}_t) - \log \pi_\theta(\mathbf{x}_{t-1}|\mathbf{x}_t)],
        \label{eq:stepwise_kl_logprob}
    \end{equation}
    where the expectation is taken over samples generated with the previous policy $\pi_{k-1}$. The full SAGE-GRPO objective that combines the GRPO policy loss, temporal equalization, and Dual KL regularization is provided in Appendix~\ref{app:full_objective}.

\section{Experiments}
\label{sec:experiments}

\subsection{Experimental Setup}
\noindent \textbf{Implementation Details.}
We conduct all experiments on HunyuanVideo 1.5~\cite{hunyuanvideo} with per-GPU batch size $2$ and $4$ gradient accumulation steps (effective batch size $8$).
Each video contains $81$ frames, and we apply GRPO updates every $20$ sampling steps along the diffusion trajectory.
Following~\cite{flowgrpo}, we use VideoAlign~\cite{videoalign} as the reward oracle, evaluating Visual Quality (VQ), Motion Quality (MQ), and Text Alignment (TA), with overall reward $R = w_{vq}S_{vq} + w_{mq}S_{mq} + w_{ta}S_{ta}$.
We compare SAGE-GRPO against DanceGRPO~\cite{dancegrpo}, FlowGRPO~\cite{flowgrpo}, and CPS~\cite{cps}.
The KL regularization weight is scheduled in $\lambda_{KL}\in[10^{-7},10^{-5}]$ according to Appendix~\ref{app:full_objective}.

\subsection{Main Results}
\label{sub:main_results}

We consider two reward configurations (Table~\ref{tab:main_results}): \textbf{averaged} $(w_{vq}=1.0, w_{mq}=1.0, w_{ta}=1.0)$ and \textbf{alignment-focused} $(w_{vq}=0.5, w_{mq}=0.5 w_{ta}=1.0)$.
All rewards use the original VideoAlign model as a frozen evaluator (no reward-model fine-tuning), which ensures consistent evaluation across methods.
Since current video GRPO baselines are implemented with substantial differences in engineering optimizations, directly reusing them would confound algorithmic effects with infrastructure choices. 
To obtain a fair comparison, we implement a unified training framework on HunyuanVideo1.5 with shared infrastructure across all methods and vary only the GRPO algorithm itself.

Under the averaged-reward setting that matches Longcat-Video~\cite{longcatvideo}, adding KL regularization typically improves visual performance but yields worse reward behavior, which we attribute to reward hacking in the reward model as discussed in previous work~\cite{selfpacedgrpo}.
We compare previous methods and SAGE-GRPO under both averaged and alignment-focused rewards, and evaluate variants with and without KL regularization, as summarized in Table~\ref{tab:main_results}.
We further study how placing more weight on semantic alignment can reduce reward hacking artifacts.
In the alignment-focused setting (Setting B), SAGE-GRPO with Dual Moving KL achieves the best Overall, VQ, MQ, and CLIPScore while remaining close to the best TA, and overall Table~\ref{tab:main_results} suggests that emphasizing alignment provides a more reliable optimization target and yields more stable gains in both reward and visual metrics.

\subsection{Qualitative Analysis}
\label{sub:qualitative}

We provide qualitative examples that complement the quantitative trends.
Figure~\ref{fig:qualitative_baseline} highlights the improvement in coherence, photorealism, and semantic alignment over baselines, especially for prompts that require precise object interactions and long-range motion.
Additional visual comparisons demonstrating superior alignment with emotional descriptions in text prompts are presented in Appendix Figure~\ref{fig:qualitative_compare}.

\subsection{User Study}
\label{sub:user_study}

To corroborate our automatic metrics, we conducted a user preference study with 29 evaluators on 32 prompts, comparing SAGE-GRPO with baselines (all at iter 100, sampling step 40, Setting B) across Visual Quality, Motion Quality, and Semantic Alignment.
Table~\ref{tab:user_study} reports the pairwise win rates of SAGE-GRPO against each baseline. 

\begin{table}[h]
    \centering
    \caption{\textbf{User Preference Study.} Win rates of SAGE-GRPO against baselines. Results indicate a strong human preference for our method, especially in Motion Quality, confirming that automatic metrics align with perceptual quality.}
    \label{tab:user_study}
    \resizebox{0.9\linewidth}{!}{
    \begin{tabular}{l|ccc}
        \toprule
        \textbf{SAGE-GRPO vs.} & \textbf{Visual Quality} & \textbf{Motion Quality} & \textbf{Semantic Alignment} \\
        \midrule
        DanceGRPO & 85.9\% & 75.8\% & 79.2\% \\
        FlowGRPO & 83.8\% & 79.2\% & 71.9\% \\
        CPS & 80.2\% & 70.8\% & 67.9\% \\
        \bottomrule
    \end{tabular}
    }
\end{table}

\subsection{Ablation Studies}

\subsubsection{Impact of Temporal Gradient Equalizer}

To evaluate the effectiveness of the Temporal Gradient Equalizer in Section~\ref{sec:method_micro}, we compare training dynamics with and without per-timestep balancing across three SDE formulations and CPS. 
Figure~\ref{fig:balance_ablation_main} shows the overall VideoAlign reward curves for baselines and our method.

\subsubsection{KL Strategy Ablation}

\begin{figure}[t]
\centering
\begin{subfigure}[b]{0.48\columnwidth}
\centering
\includegraphics[width=\linewidth]{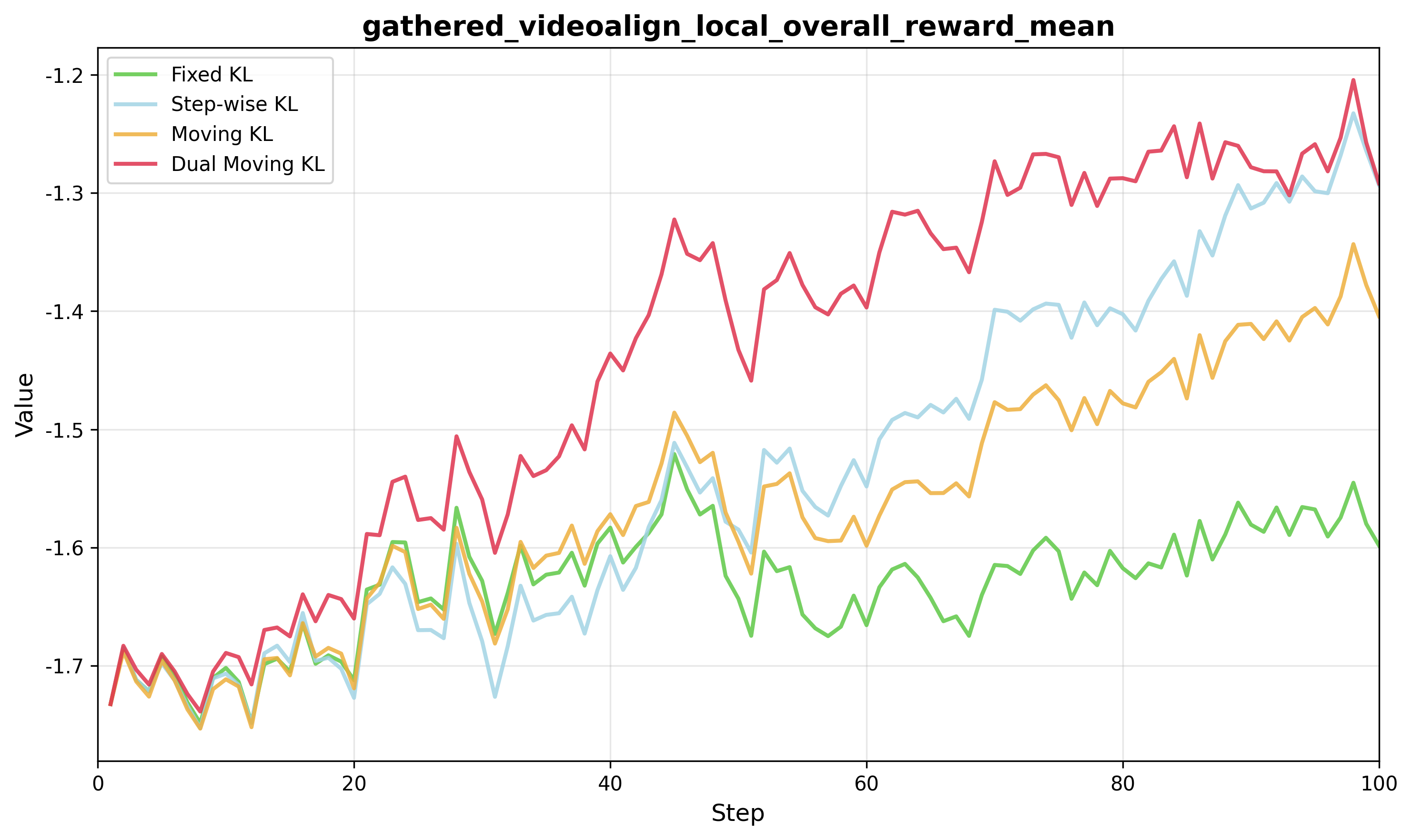}
\caption{Mean reward}
\label{fig:kl_ablation_mean}
\end{subfigure}
\hfill
\begin{subfigure}[b]{0.48\columnwidth}
\centering
\includegraphics[width=\linewidth]{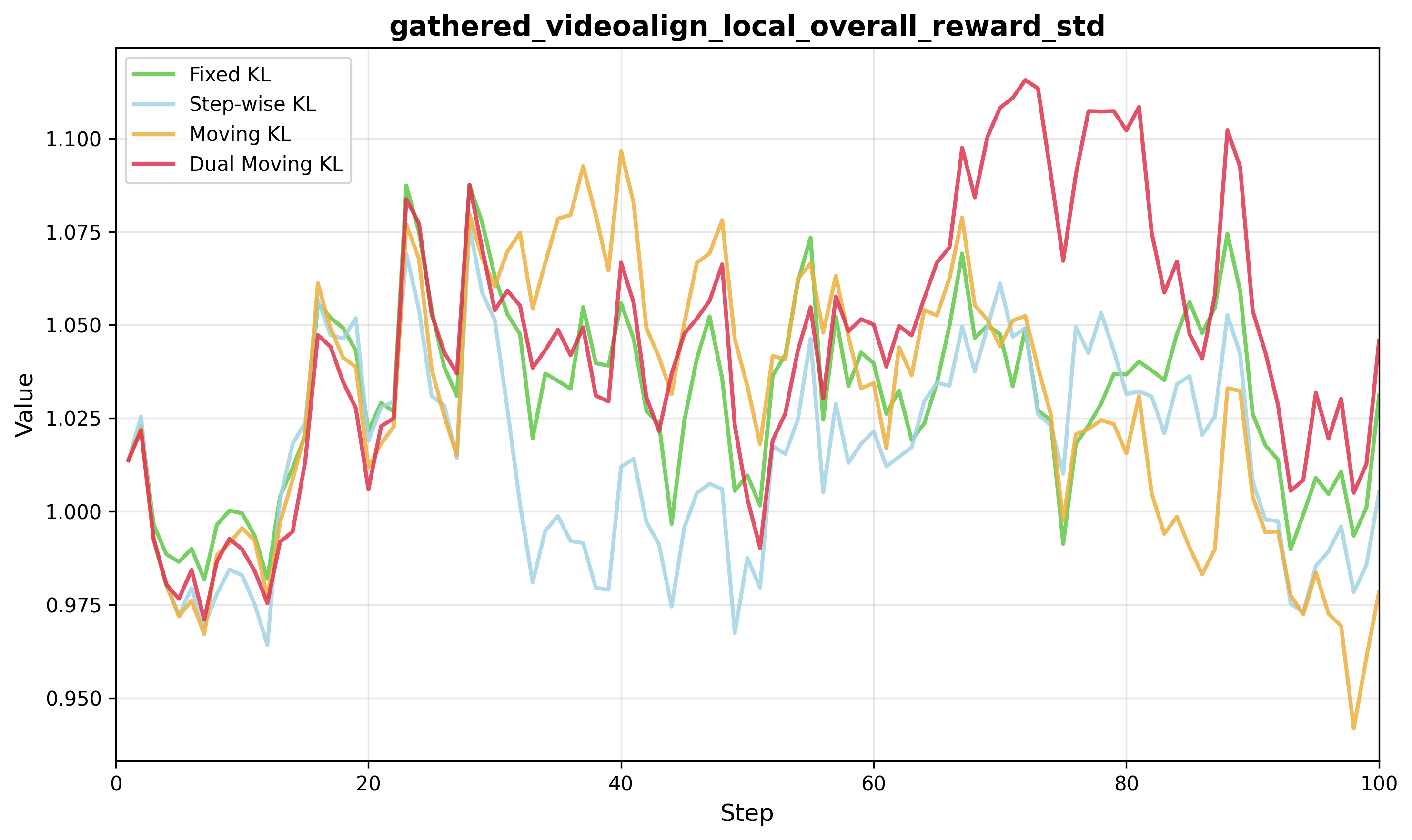}
\caption{Std (exploration)}
\label{fig:kl_ablation_std}
\end{subfigure}
\caption{\textbf{KL strategy ablation.}
(a) Dual Moving KL achieves the highest and most stable reward, supporting the position-velocity control interpretation (Equation~\eqref{eq:dual_kl}).
(b) Moving KL attains high exploration in early training steps but the exploration level drops in later stages. Dual Moving KL maintains a higher and more stable exploration level throughout training.}
\label{fig:kl_ablation}
\end{figure}

We next study the effect of different KL strategies introduced in Section~\ref{sec:method_dual_kl}. 
Figure~\ref{fig:kl_ablation} reports both the mean reward and standard deviation for four KL strategies, with qualitative comparisons in Appendix Figures~\ref{fig:kl_ablation_qualitative_1} and~\ref{fig:kl_ablation_qualitative_2}.

Figure~\ref{fig:kl_ablation_mean} shows that Dual Moving KL consistently outperforms other variants in both convergence speed and final reward while avoiding the collapse observed in aggressive step-wise updates.
Figure~\ref{fig:kl_ablation_std} shows that Moving KL explores quickly initially but exploration falls off; Dual Moving KL maintains higher exploration stably, validating the position-velocity controller interpretation in Equation~\eqref{eq:dual_kl}.

\subsubsection{KL Weight Sensitivity}

We compare three KL weight schedules: fixed $10^{-5}$, two-stage $10^{-7} \rightarrow 10^{-5}$, and milder $10^{-7} \rightarrow 10^{-6}$.
Figure~\ref{fig:kl_weight_ablation} shows that the two-stage schedule yields higher rewards and smoother trajectories across VQ, MQ, and TA, consistent with gradually increasing $\lambda_{KL}$ to tighten the trust region.
Implementation details are in Appendix~\ref{app:adaptive_kl}.

\section{Conclusion}
\label{sec:conclusion}

We presented \textbf{SAGE-GRPO}, a manifold-aware GRPO framework for stable reinforcement learning for video generation.
The core challenge is to design exploration strategies that respect the manifold structure, where each exploration step stays within the vicinity of the manifold rather than drifting into high-noise regions.
At the micro-level, we derive a Precise Manifold-Aware SDE that keeps exploration noise closer to the flow trajectory, and introduce a Gradient Norm Equalizer that normalizes optimization pressure across timesteps.
At the macro-level, we propose a Dual Trust Region mechanism combining position and velocity control to reduce off-manifold local optima while enabling sustained plasticity.
Experiments on HunyuanVideo1.5 with VideoAlign reward show consistent improvements over strong baselines and validate the contribution of each component through ablations.

\section*{Impact Statement}

This paper presents a method for more stable reinforcement learning alignment of text-to-video generation models.
By improving temporal consistency and text alignment under a fixed reward model, our work may strengthen creative tools, scientific communication, and educational content that rely on controllable video synthesis.
At the same time, stronger video generation systems can exacerbate existing concerns about misinformation, deepfakes, biased or harmful content, and the computational cost of large-scale training and sampling.
Our experiments are conducted in a research setting on an existing model and evaluator, and our user study involves 29 voluntary evaluators rating 32 prompts comparing SAGE-GRPO against baselines in terms of visual quality, motion quality, and semantic alignment; there is no collection of personal data, but any future deployment should include safeguards such as content moderation, dataset auditing, and human oversight to reduce these risks.
\nocite{langley00}

\bibliography{example_paper}
\bibliographystyle{icml2026}

\newpage
\appendix
\onecolumn
\section{Appendix}
\label{sec:appendix}

\subsection{Derivation of Manifold-Aware SDE Variance}
\label{app:sde_derivation}

\paragraph{Derivation}

To enable stochastic exploration for GRPO, we need to convert the deterministic Rectified Flow ODE into a stochastic differential equation (SDE) that preserves the marginal probability distribution at each timestep.
Recall the general form of a marginal-preserving SDE for flow matching:
\begin{equation}
    \mathrm{d}\mathbf{z}_t
= \Big(\mathbf{v}_\theta(\mathbf x_t,t) - \tfrac{1}{2}\varepsilon_t^2 \mathbf s_\theta(\mathbf x_t)\Big)\,\mathrm{d}t
+ \varepsilon_t\,\mathrm{d}\mathbf{w}_t,
\end{equation}
where $\varepsilon_t$ is the diffusion coefficient (a function of $t$), $\mathbf{w}_t$ is a Brownian motion, and $\mathbf{s}_\theta(\mathbf{x}_t) \approx -(\mathbf{x}_t - \hat{\mathbf{x}}_0)/\sigma_t^2$ is the score function estimate.
The It\^{o} correction term $\tfrac{1}{2}\varepsilon_t^2 \mathbf{s}_\theta(\mathbf{x}_t)$ ensures that the SDE preserves the same marginal distribution as the deterministic ODE.

To discretize this SDE, we assume that $\mathbf{v}_\theta(\mathbf{x}_t, t)$ and $\mathbf{s}_\theta(\mathbf{x}_t)$ remain approximately constant during the interval $[\sigma_{t+1}, \sigma_t]$, where $\sigma_t$ is the noise level at timestep $t$.
The key challenge is to compute the integrated variance $\Sigma_t$ for the stochastic term.
We define:
\begin{equation}
    \Sigma_t \coloneq \int_{\sigma_{t+1}}^{\sigma_t} \varepsilon_s^2 \,\mathrm{d}s.
\end{equation}
For Rectified Flow, we choose $\varepsilon_t = \eta \sqrt{\frac{\sigma_t}{1-\sigma_t}}$ to match the geometric structure of the flow trajectory, where $\eta$ is the exploration scaling factor.
Substituting this form and integrating:
\begin{align}
    \Sigma_t &= \int_{\sigma_{t+1}}^{\sigma_t} \eta^2 \frac{\sigma_s}{1-\sigma_s} \,\mathrm{d}s \\
    &= \eta^2 \int_{\sigma_{t+1}}^{\sigma_t} \left( \frac{1}{1-\sigma_s} - 1 \right) \,\mathrm{d}s \\
    &= \eta^2 \left[ -\log(1-\sigma_s) - \sigma_s \right]_{\sigma_{t+1}}^{\sigma_t} \\
    &= \eta^2 \left[ -(\sigma_t - \sigma_{t+1}) + \log\left(\frac{1-\sigma_{t+1}}{1-\sigma_t}\right) \right].
\end{align}
Taking the square root, we obtain the noise standard deviation:
\begin{equation}
    \Sigma_t^{1/2} = \eta \sqrt{-(\sigma_t - \sigma_{t+1}) + \log\left(\frac{1-\sigma_{t+1}}{1-\sigma_t}\right)}.
    \label{eq:precise_std}
\end{equation}
The logarithmic term $\log((1-\sigma_{t+1})/(1-\sigma_t))$ accounts for the geometric contraction of the signal coefficient $(1-\sigma_t)$, which linear approximations fail to capture.

Applying Euler-Maruyama discretization with timestep $\Delta t = \sigma_t - \sigma_{t+1}$, the discretized SDE becomes:
\begin{equation}
    \mathbf{x}_{t+\Delta t} = \mathbf{x}_t + \mathbf{v}_\theta(\mathbf{x}_t, t)\Delta t + \frac{\Sigma_t}{2} \mathbf{s}_\theta(\mathbf{x}_t) + \Sigma_t^{1/2} \bm{\epsilon},
    \label{eq:discretized_sde}
\end{equation}
where $\bm{\epsilon} \sim \mathcal{N}(\mathbf{0}, \mathbf{I})$ is the injected stochasticity.
Note that $\Sigma_t$ is already the integrated variance over the interval $[\sigma_{t+1}, \sigma_t]$, so the stochastic term uses $\Sigma_t^{1/2}$ directly without an additional $\sqrt{\Delta t}$ factor.

\paragraph{Problem Formulation.}
Let the noise level at timestep $t$ be $\sigma_t$. In a Rectified Flow setting, the trajectory connects pure noise ($\sigma=1$) to data ($\sigma=0$). We aim to find the precise variance $\Sigma_t$ required for the stochastic step such that the marginal distribution is preserved up to the second order.

Let $\Delta \sigma = \sigma_t - \sigma_{t+1} > 0$. We analyze the terms inside the square root of our proposed Eq.~\ref{eq:precise_sde}. Let $V_t$ denote the variance term:
\begin{equation}
    V_t = -(\sigma_t - \sigma_{t+1}) + \log\left(\frac{1-\sigma_{t+1}}{1-\sigma_t}\right)
\end{equation}

\paragraph{Taylor Expansion Analysis.}
First, we express the logarithmic term using $\Delta \sigma$:
\begin{equation}
    \log\left(\frac{1-\sigma_{t+1}}{1-\sigma_t}\right) = \log\left(\frac{1-(\sigma_t - \Delta \sigma)}{1-\sigma_t}\right) = \log\left(1 + \frac{\Delta \sigma}{1-\sigma_t}\right)
\end{equation}
Let $x = \frac{\Delta \sigma}{1-\sigma_t}$. Since step sizes are small, $|x| < 1$. We apply the Taylor expansion $\log(1+x) \approx x - \frac{x^2}{2} + \mathcal{O}(x^3)$:
\begin{equation}
    \log\left(1 + \frac{\Delta \sigma}{1-\sigma_t}\right) \approx \frac{\Delta \sigma}{1-\sigma_t} - \frac{1}{2} \left(\frac{\Delta \sigma}{1-\sigma_t}\right)^2
\end{equation}

Substituting this back into $V_t$:
\begin{align}
    V_t &\approx -\Delta \sigma + \left( \frac{\Delta \sigma}{1-\sigma_t} - \frac{1}{2} \frac{\Delta \sigma^2}{(1-\sigma_t)^2} \right) \\
    &= \Delta \sigma \left( \frac{1}{1-\sigma_t} - 1 \right) - \frac{1}{2} \frac{\Delta \sigma^2}{(1-\sigma_t)^2} \\
    &= \Delta \sigma \left( \frac{\sigma_t}{1-\sigma_t} \right) - \mathcal{O}(\Delta \sigma^2)
\end{align}
The leading term $\Delta \sigma \frac{\sigma_t}{1-\sigma_t}$ represents the ideal variance scaling for a geometric schedule, which linear approximations fail to capture.

\subsection{Standard Deviation Comparison: Ours vs.\ FlowGRPO}
\label{app:std_comparison}

\begin{figure*}[t]
\centering
\begin{subfigure}[b]{0.32\textwidth}
\centering
\includegraphics[width=\linewidth]{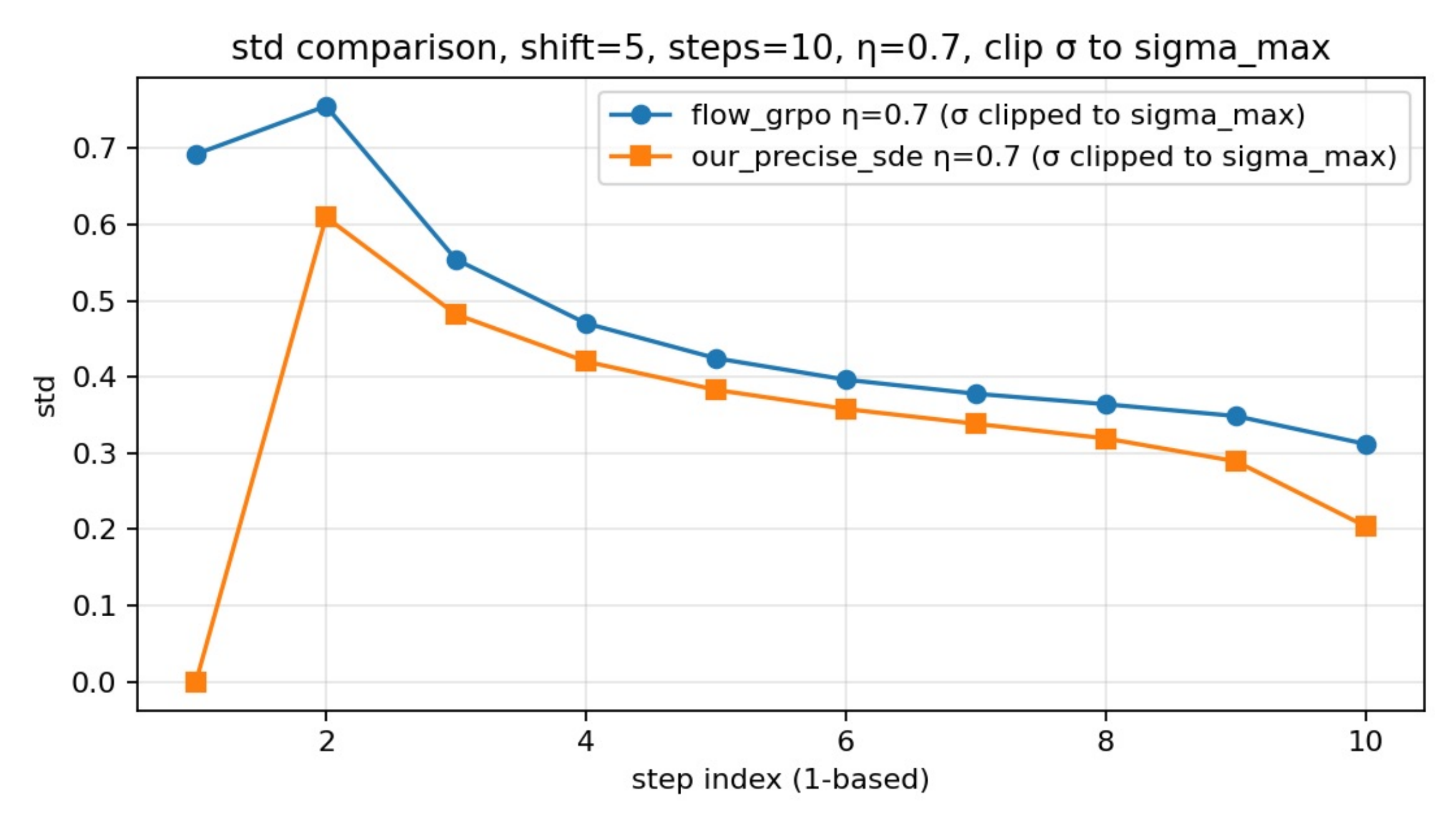}
\caption{Both using FlowGRPO's $\sigma$ schedule.}
\label{fig:app_std_a}
\end{subfigure}
\hfill
\begin{subfigure}[b]{0.32\textwidth}
\centering
\includegraphics[width=\linewidth]{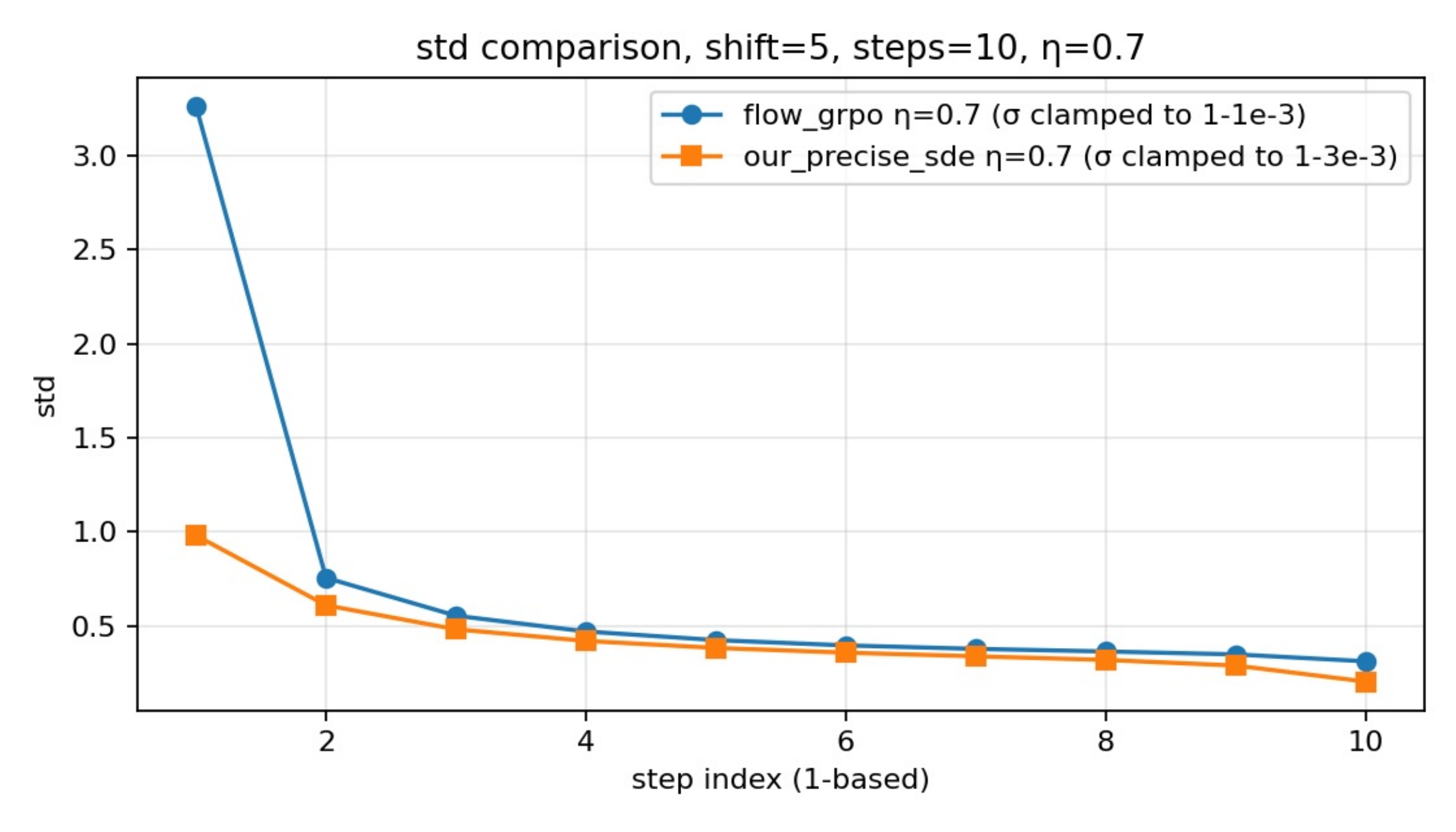}
\caption{Both clamped at $1-3 \times 10^{-3}$.}
\label{fig:app_std_b}
\end{subfigure}
\hfill
\begin{subfigure}[b]{0.32\textwidth}
\centering
\includegraphics[width=\linewidth]{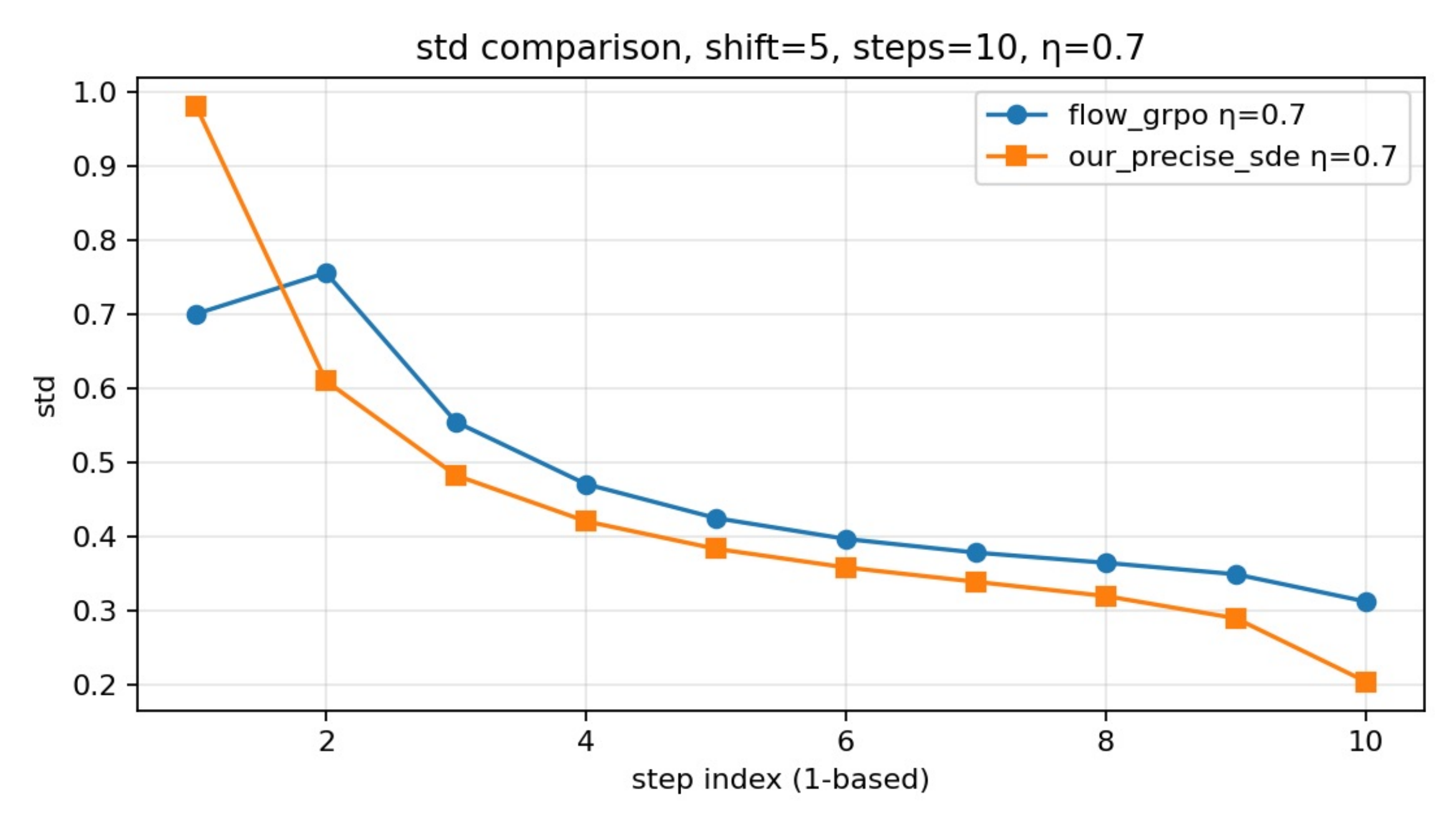}
\caption{Each using default implementation.}
\label{fig:app_std_c}
\end{subfigure}
\caption{\textbf{Step-wise std comparison: our precise SDE vs.\ FlowGRPO.}
(a) When both use FlowGRPO's $\sigma$ schedule, our integral-based formulation yields near-zero std at the first step due to equal endpoints. (b) When both are clamped at $(1-\sigma) \geq 3 \times 10^{-3}$, FlowGRPO explodes at step 1, while ours remains stable. (c) Under default implementations, ours maintains lower std across most steps. This supports that we remove ineffective high-frequency noise and explore more precisely along the manifold (Section~\ref{sec:method_micro}).}
\label{fig:app_std_comparison}
\end{figure*}

Figure~\ref{fig:app_std_comparison} compares the noise standard deviation per step between our precise SDE and FlowGRPO under three parameterization regimes to understand how different noise handling strategies affect exploration behavior.

\noindent \textbf{Regime (a): Both methods using FlowGRPO's $\sigma$ schedule.}
When both methods use FlowGRPO's default $\sigma$ schedule (where $\sigma_t$ is set to $\sigma_{\max}$ at early steps), our precise SDE exhibits near-zero standard deviation at the first step.
This occurs because our method computes noise variance via integration: $\Sigma_t = \int_{\sigma_{t+1}}^{\sigma_t} \varepsilon_s^2 \,\mathrm{d}s$.
When both endpoints are equal ($\sigma_t = \sigma_{t+1} = \sigma_{\max}$), the integration interval collapses, and the logarithmic term $\log((1-\sigma_{t+1})/(1-\sigma_t))$ evaluates to zero, yielding $\Sigma_t \approx 0$.
This demonstrates that our integral-based formulation is sensitive to the $\sigma$ schedule and requires proper boundary handling.

\noindent \textbf{Regime (b): Both methods using aggressive clamping at $1-3 \times 10^{-3}$.}
When both methods apply the same clamping threshold $(1-\sigma) \geq 3 \times 10^{-3}$ (equivalently, $\sigma \leq 1-3 \times 10^{-3}$), FlowGRPO exhibits explosive behavior at the first step, with standard deviation reaching values around $3.0$.
This instability arises because FlowGRPO's noise computation involves a ratio $\sigma/(1-\sigma)$; when $(1-\sigma)$ is clamped to a small constant while $\sigma$ remains large, the denominator becomes artificially small, causing the ratio to explode.
In contrast, our precise SDE maintains stable and controlled standard deviation throughout, starting around $1.0$ and decaying smoothly, demonstrating that our manifold-aware formulation inherently handles low-noise regimes more robustly.

\noindent \textbf{Regime (c): Each method using its default implementation.}
Under their respective default configurations, FlowGRPO uses its standard $\sigma$ schedule, while our method applies clamping at $(1-\sigma) \geq 3 \times 10^{-3}$.
Our method maintains a lower standard deviation than FlowGRPO across most of the diffusion trajectory, particularly in later steps.
This demonstrates that our precise SDE effectively reduces injected noise magnitude, leading to more refined exploration along the data manifold.

Across all three regimes, our method consistently achieves smaller or more stable standard deviation than FlowGRPO.
This supports the main-figure narrative (Figure~\ref{fig:teaser}): we remove unnecessary high-frequency noise energy in high-noise regions, enabling more precise exploration that stays closer to the data manifold.
This behavior aligns with the micro-level exploration design of our SDE in Section~\ref{sec:method_micro}.

\begin{figure*}[t]
    \centering
    \includegraphics[width=\textwidth]{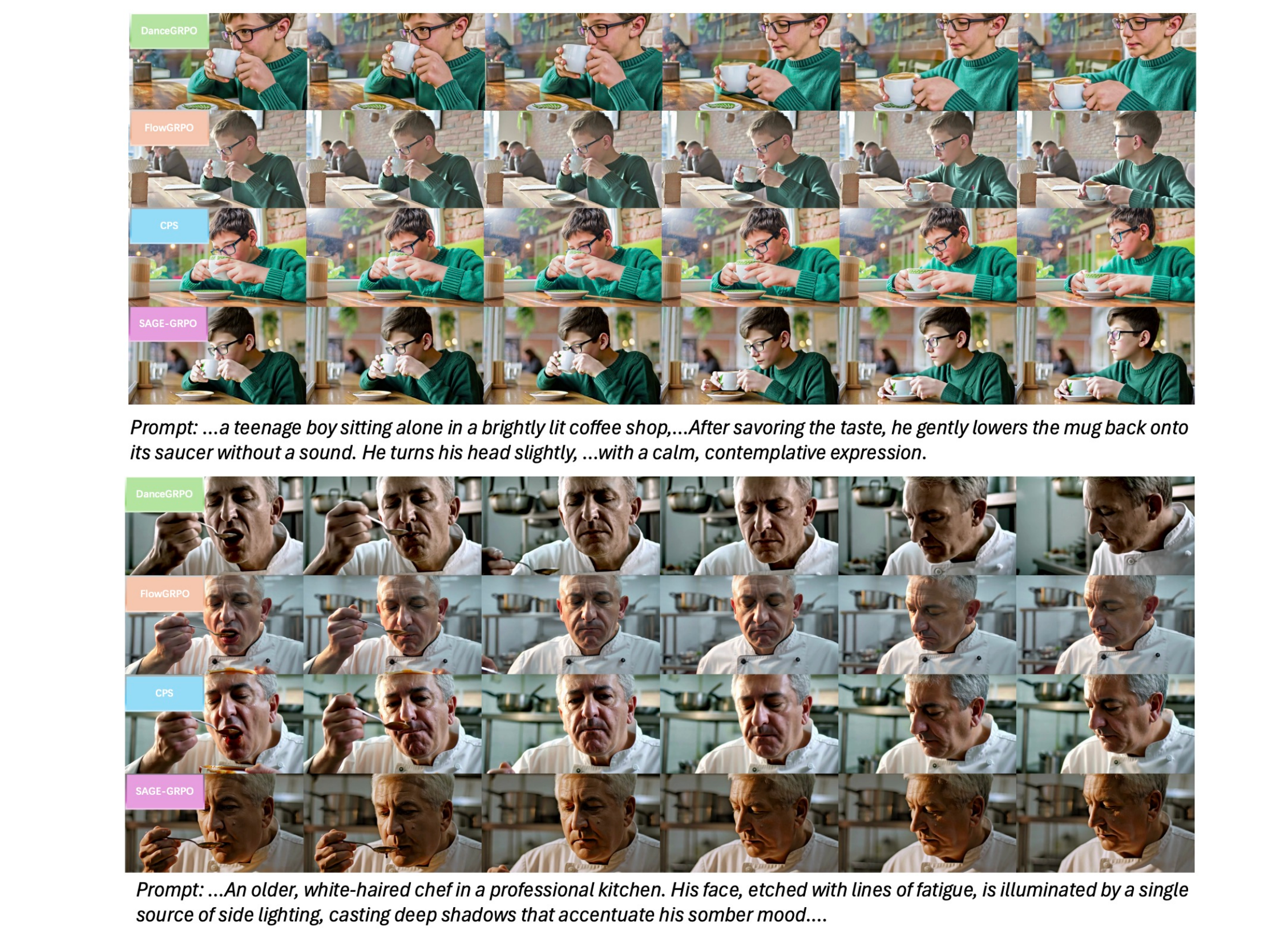}
    \caption{\textbf{Qualitative comparison highlighting emotional alignment.}
    Two prompts illustrate SAGE-GRPO's ability to better align with emotional descriptions: (Top) A teenage boy in a coffee shop, where SAGE-GRPO captures the "calm, contemplative expression" and gentle motion of lowering the mug, while baselines show neutral expressions and abrupt movements. (Bottom) An older chef in a kitchen, where SAGE-GRPO consistently renders the "lines of fatigue" and "somber mood" through deep side-lighting shadows, while baselines fail to convey the intended emotional depth. Our manifold-aware exploration enables precise alignment with subtle emotional and action cues.}
    \label{fig:qualitative_compare}
\end{figure*}

\subsection{Theoretical Gradient Norm Analysis}
\label{app:grad_norm}

Here we derive the relationship between the gradient norm and the noise schedule.
For a Gaussian policy $\pi(\mathbf{x}_{t-1}|\mathbf{x}_t) = \mathcal{N}(\mu_\theta, \Sigma_t \mathbf{I})$, the gradient of the log-probability with respect to the drift parameter $\mu_\theta$ is:
\begin{equation}
    \nabla_{\mu} \log \pi = \frac{\mathbf{x}_{sample} - \mu_\theta}{\Sigma_t}
\end{equation}
Since $\mathbf{x}_{sample} \sim \mathcal{N}(\mu_\theta, \Sigma_t)$, the expected norm is proportional to the standard deviation of the noise:
\begin{equation}
    \mathbb{E}[\|\nabla_{\mu} \log \pi\|] \propto \frac{\sqrt{\Sigma_t}}{\Sigma_t} = \frac{1}{\sqrt{\Sigma_t}}
\end{equation}
Given our derived Manifold-Aware variance $\Sigma_t \approx \eta^2 \Delta \sigma \frac{\sigma_t}{1-\sigma_t}$, the gradient norm scales as:
\begin{equation}
    \|\nabla\| \propto \sqrt{\frac{1-\sigma_t}{\sigma_t \Delta \sigma}}
\end{equation}
This confirms that as $\sigma_t \to 0$ (low noise), the gradient norm explodes, necessitating our proposed Gradient Equalizer.

\subsection{GRPO Reward and Advantage Details}
\label{app:grpo_details}

Here we provide the implementation-aligned definitions of reward composition and the group-normalized advantage used in Equation~\eqref{eq:grpo_objective}.
Following VideoAlign~\cite{videoalign}, we construct a composite reward for a generated video $\mathbf{x}_0$:
\begin{equation}
    R(\mathbf{x}_0)=w_{vq}S_{vq}(\mathbf{x}_0)+w_{mq}S_{mq}(\mathbf{x}_0)+w_{ta}S_{ta}(\mathbf{x}_0),
\end{equation}
where $S_{vq}$, $S_{mq}$, and $S_{ta}$ score visual quality, motion quality, and text alignment, and $w_{vq},w_{mq},w_{ta}$ are fixed scalar weights.

Given a prompt $\mathbf{c}$, GRPO samples a group of $G$ rollouts $\{\mathbf{x}_0^{(i)}\}_{i=1}^G$ and computes rewards $r_i=R(\mathbf{x}_0^{(i)})$.
We use the group mean and standard deviation as a baseline:
\begin{equation}
    \mu_R=\frac{1}{G}\sum_{j=1}^G r_j,\qquad
    \sigma_R=\sqrt{\frac{1}{G}\sum_{j=1}^G (r_j-\mu_R)^2},
\end{equation}
and define the normalized advantage:
\begin{equation}
    A_i=\frac{r_i-\mu_R}{\sigma_R+\epsilon},
\end{equation}
where $\epsilon$ is a small constant for numerical stability.

\subsection{Temporal Gradient Equalizer: Derivation of $\mathcal{N}_t$}
\label{app:equalizer_details}

We outline how to obtain a per-timestep gradient scale proxy $\mathcal{N}_t$ that is compatible with the SDE transition used in Section~\ref{sec:method_micro}.
Consider a Gaussian transition $\pi(\mathbf{x}_{t-1}\mid\mathbf{x}_t)=\mathcal{N}(\bm{\mu}_\theta, \Sigma_t\mathbf{I})$ parameterized through the network output (e.g., velocity/denoiser prediction) and a noise variance $\Sigma_t$ determined by the chosen SDE.
The log-probability gradient with respect to the mean parameter satisfies:
\begin{equation}
    \nabla_{\bm{\mu}}\log\pi=\frac{\mathbf{x}_{sample}-\bm{\mu}_\theta}{\Sigma_t}.
\end{equation}
Since $\mathbf{x}_{sample}-\bm{\mu}_\theta\sim\mathcal{N}(\mathbf{0},\Sigma_t\mathbf{I})$, its magnitude is $\mathcal{O}(\Sigma_t^{1/2})$ in expectation, yielding the inverse relationship:
\begin{equation}
    \mathbb{E}\big[\|\nabla_{\bm{\mu}}\log\pi\|\big]\propto \frac{1}{\Sigma_t^{1/2}}.
\end{equation}

In practice, the network does not directly parameterize $\bm{\mu}_\theta$; instead, $\bm{\mu}_\theta$ is obtained by composing the network prediction with the SDE/solver update rule, introducing an additional sensitivity factor.
Let $\lambda_t$ denote the scalar sensitivity from the solver mapping (details depend on the SDE type and discretization).
We use the proxy:
\begin{equation}
    \mathcal{N}_t=\frac{\lambda_t}{\Sigma_t^{1/2}},
\end{equation}
and define the Temporal Gradient Equalizer (Equation~\eqref{eq:grad_equalizer}) as a robust normalization:
\begin{equation}
    S_t=\frac{\mathrm{Median}(\{\mathcal{N}_\tau\}_{\tau=1}^T)}{\mathcal{N}_t+\epsilon}.
\end{equation}
This produces approximately uniform gradient scales across timesteps, aligning with the empirical observation in Figure~\ref{fig:grad_norm_analysis} and the training-curve improvement in Figure~\ref{fig:balance_ablation_main}.

\subsection{SAGE-GRPO Objective and Adaptive KL Weighting}
\label{app:full_objective}

We provide the complete objective used in SAGE-GRPO, combining GRPO, the Temporal Gradient Equalizer, and Dual KL regularization, together with a principled schedule for the overall KL coefficient.
At each optimization step, we sample a group of $G$ rollouts and compute advantages $\{A_i\}_{i=1}^G$ as in Appendix~\ref{app:grpo_details}.

\noindent \textbf{Dual KL regularizer.}
We use two reference policies to implement a position--velocity controller in policy space (Section~\ref{sec:method_dual_kl}). The regularizer is
\begin{equation}
    \mathcal{L}_{KL}=\beta_{pos}\cdot D_{KL}(\pi_\theta\|\pi_{ref\_N})+\beta_{vel}\cdot D_{KL}(\pi_\theta\|\pi_{k-1}),
\end{equation}
where $\pi_{k-1}$ is the previous policy and $\pi_{ref\_N}$ is a periodically refreshed anchor.
The term $D_{KL}(\pi_\theta\|\pi_{k-1})$ constrains the \emph{instantaneous update} (velocity control), while $D_{KL}(\pi_\theta\|\pi_{ref\_N})$ constrains the \emph{cumulative displacement} from the anchor (position control).
This separation is important for long-horizon training: velocity-only constraints can still accumulate drift, whereas a single fixed anchor can be overly restrictive.

\noindent \textbf{Adaptive KL weighting.}
We interpret the overall KL coefficient $\lambda_{KL}$ as a Lagrange multiplier associated with a trust-region constraint $\mathbb{E}[D_{KL}(\pi_\theta\|\pi_{ref})]\leq \delta$.
Instead of fixing $\lambda_{KL}$, we adapt it online so that the realized KL remains close to a target scale, analogous in spirit to adaptive behavior regularization in AWAC~\cite{AWAC}, where a temperature parameter is adapted from advantage statistics.
\label{app:adaptive_kl}

\noindent \emph{Warm-up (two-stage increase).}
Let $\lambda_{\min}=10^{-7}$ and $\lambda_{\max}=10^{-5}$ denote the minimum and maximum KL coefficients.
During the first $K=100$ optimization steps, we use a linear warm-up:
\begin{equation}
    \lambda_{KL}(k)=\lambda_{\min}+\left(\lambda_{\max}-\lambda_{\min}\right)\cdot \frac{k}{K}, \qquad k\le K,
    \label{eq:kl_warmup}
\end{equation}
which corresponds to the two-stage schedules reported in Figure~\ref{fig:kl_weight_ablation}.
This design keeps the trust region weak early to avoid underfitting, and gradually strengthens it as the policy improves.

\noindent \emph{Conservative feedback control.}
After warm-up, we apply a proportional feedback update based on the recent KL history, similar to the $P$-term of a PID controller.
Let $\bar{D}_{KL}$ be the mean of the last $H=10$ observed KL values and let $D_{target}$ be the desired KL scale.
We define the KL error $e_{KL}=\bar{D}_{KL}-D_{target}$ and update
\begin{equation}
    \lambda_{KL}\leftarrow
    \begin{cases}
        0.9\,\lambda_{KL}, & \bar{D}_{KL} > (1+0.5)\,D_{target},\\
        1.1\,\lambda_{KL}, & \bar{D}_{KL} < (1-0.5)\,D_{target},\\
        \lambda_{KL}, & \text{otherwise},
    \end{cases}
    \qquad \lambda_{KL}\in[\lambda_{\min},\lambda_{\max}],
    \label{eq:kl_controller}
\end{equation}
where clipping enforces the same bounds as the warm-up stage.
Intuitively, if the empirical KL is much larger than $D_{target}$, the controller reduces $\lambda_{KL}$ to relax the constraint; if it is much smaller, the controller increases $\lambda_{KL}$ to tighten the trust region.
This combination of warm-up and feedback control stabilizes the effective trust-region radius and explains the smooth reward trajectories observed in Figure~\ref{fig:kl_weight_ablation}.

\noindent \textbf{Full SAGE-GRPO loss.}
Combining GRPO, the Temporal Gradient Equalizer, and the adaptively weighted Dual KL regularizer yields:
\begin{equation}
    \boxed{
    \mathcal{L}_{SAGE\text{-}GRPO}(\theta)=
    -\frac{1}{G}\sum_{i=1}^G A_i\cdot\sum_{t=1}^T S_t\cdot
    \log\pi_\theta(\mathbf{x}_{t-1}^{(i)}\mid\mathbf{x}_t^{(i)},\mathbf{c})
    -\lambda_{KL}\cdot\mathcal{L}_{KL}.
    }
\end{equation}

\subsection{Additional Qualitative Results}
\label{app:qualitative}

We include additional qualitative visualizations to complement the quantitative experiments in Section~\ref{sec:experiments}.
Figure~\ref{fig:qualitative_compare} demonstrates SAGE-GRPO's superior ability to align with emotional descriptions in text prompts, capturing subtle facial expressions and mood cues that baselines often miss.

\begin{figure*}[t]
    \centering
    \includegraphics[width=\textwidth]{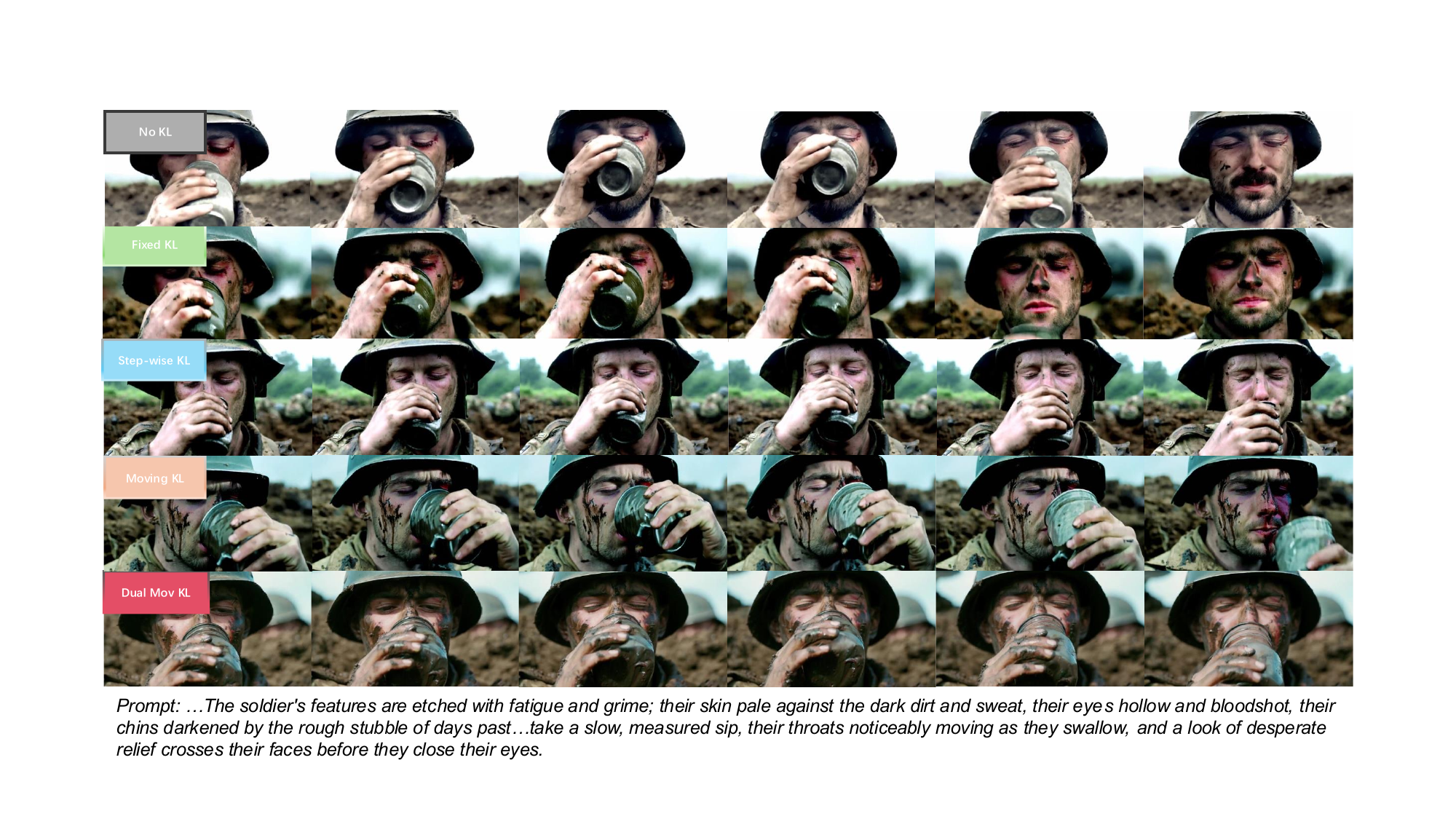}
    \caption{\textbf{KL strategy ablation: qualitative comparison (Case 1).}
    Visual comparison across different KL strategies (no KL, Fixed KL, Step-wise KL, Moving KL, Dual Moving KL) on a prompt describing a fatigued soldier. Dual Moving KL produces more realistic facial details, better dirt and grime rendering, and maintains temporal consistency across frames compared to other variants.}
    \label{fig:kl_ablation_qualitative_1}
\end{figure*}

\begin{figure*}[t]
    \centering
    \includegraphics[width=\textwidth]{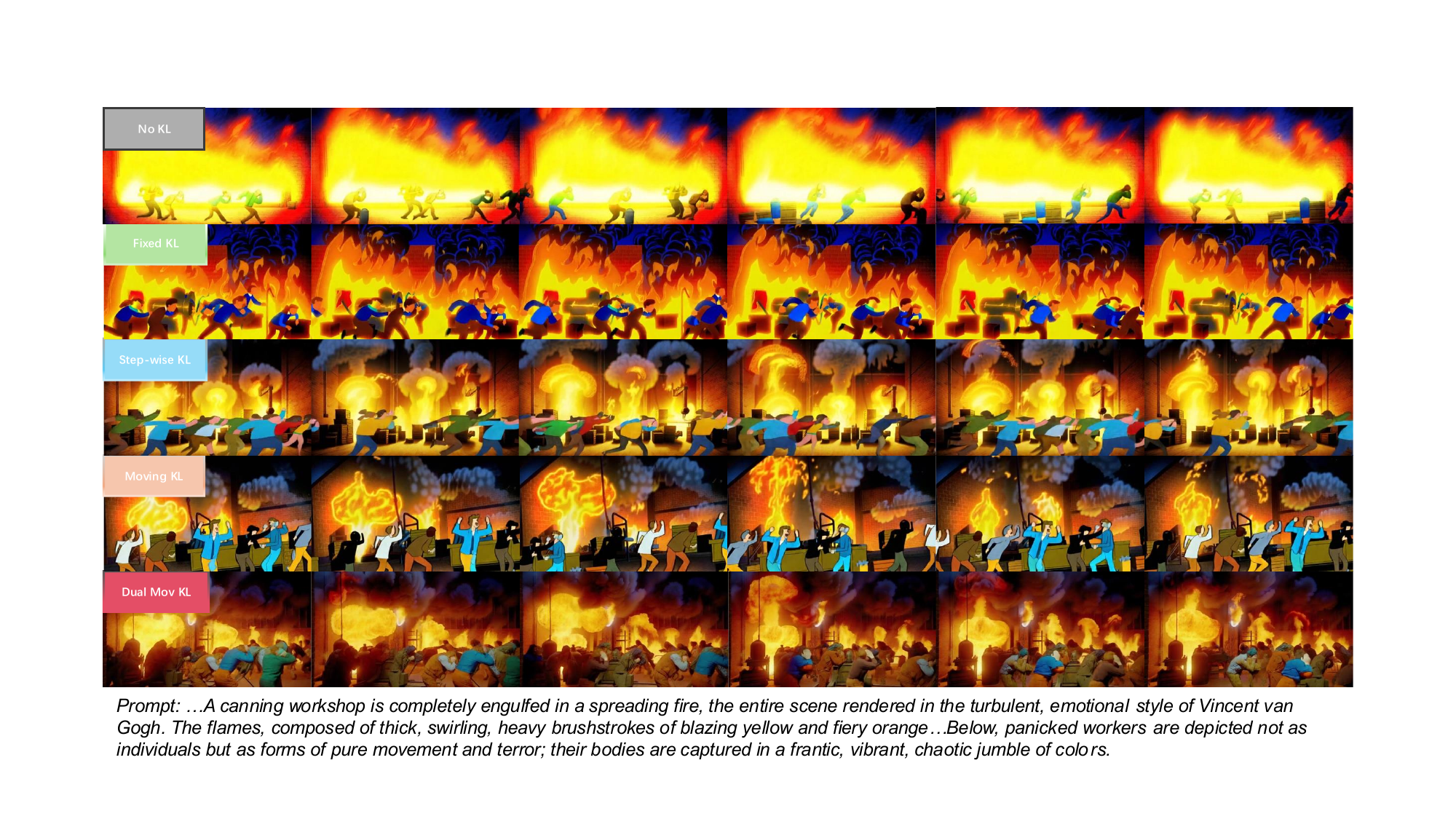}
    \caption{\textbf{KL strategy ablation: qualitative comparison (Case 2).}
    Additional visual comparison demonstrating how different KL strategies affect generation quality. Dual Moving KL consistently achieves better photorealism and alignment with prompt descriptions compared to alternatives, validating the position-velocity control mechanism discussed in Section~\ref{sec:method_dual_kl}.}
    \label{fig:kl_ablation_qualitative_2}
\end{figure*}

To further validate the effectiveness of different KL strategies discussed in Section~\ref{sec:method_dual_kl}, we provide qualitative comparisons across five variants: no KL regularization, Fixed KL (anchored to the initial model $\pi_0$), Step-wise KL (velocity control only), Moving KL (position control only), and Dual Moving KL (combining both position and velocity control).
Figures~\ref{fig:kl_ablation_qualitative_1} and~\ref{fig:kl_ablation_qualitative_2} show that Dual Moving KL consistently produces more realistic details, better temporal consistency, and stronger alignment with prompt descriptions compared to other KL strategies, which aligns with the quantitative findings in Section~\ref{sub:main_results}.

\end{document}